\documentclass[review]{elsarticle}
\usepackage[flushleft]{threeparttable}
\usepackage{lineno,hyperref}
\usepackage{graphicx}
\usepackage{amsmath,amssymb} 
\usepackage{color}
\usepackage{times}
\usepackage{epsfig}
\usepackage{mathrsfs}
\usepackage{fancybox}
\usepackage{algorithm}
\usepackage{algpseudocode}
\usepackage{pifont}
\usepackage{booktabs}
\usepackage{verbatim}  
\usepackage{cleveref}
\usepackage[utf8]{inputenc}
\usepackage[export]{adjustbox}
\usepackage{multirow}
\usepackage{array}
\usepackage{subfig}
\usepackage{blindtext}

\journal{Pattern Recognition}

\begin{document}

\begin{frontmatter}

\title{From Superpixel to Human Shape Modelling for Carried Object Detection}



\author[mymainaddress]{Farnoosh Ghadiri\corref{mycorrespondingauthor}}
\cortext[mycorrespondingauthor]{Corresponding author: Farnoosh ghadiri}
\ead{farnoosh.ghadiri.1@ulaval.ca}

\author[mymainaddress]{Robert Bergevin}
\ead{bergevin@gel.ulaval.ca}

\author[mysecondaryaddress]{Guillaume-Alexandre Bilodeau}
\ead{gabilodeau@polymtl.ca}

\address[mymainaddress]{LVSN-REPARTI, Universit\'{e} Laval,  Qu\'{e}bec, Canada}
\address[mysecondaryaddress]{ LITIV lab,  Polytechnique Montr\'{e}al,  Montr\'{e}al, Canada}

\begin{abstract}
Detecting carried objects is one of the requirements for developing systems to reason about activities involving people and objects. We present an approach to detect carried objects from a single video frame with a novel method that incorporates features from multiple scales. Initially, a foreground mask in a video frame is segmented into multi-scale superpixels. Then the human-like regions in the segmented area are identified by matching a set of extracted features from superpixels against learned features in a codebook. A carried object probability map is generated using the complement of the matching probabilities of superpixels to human-like regions and background information. A group of superpixels with high carried object probability and strong edge support is then merged to obtain the shape of the carried object. We applied our method to two challenging datasets, and results show that our method is competitive with or better than the state-of-the-art.
\end{abstract}


\end{frontmatter}


\section{Introduction}

A criminal can carry a variety of objects such as tools to perform an illegal action. To stop or prevent criminal acts, an intelligent visual surveillance system is needed to recognize suspicious activities or the tools required to perform them. However, algorithms that can detect unusual events such as exchanging objects or leaving a bag behind have proven highly difficult to devise. Since detecting objects carried by people is a key ability required by most video surveillance systems, we focused on this topic in this work.

\begin{figure}[p]
\captionsetup[subfigure]{labelformat=empty}
\centering
\subfloat[]{\includegraphics[width= 1.6cm,height=3cm]{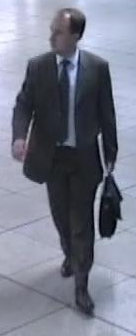}}
\hspace{0.3cm}
\subfloat[]{\includegraphics[width= 1.6cm,height=3cm]{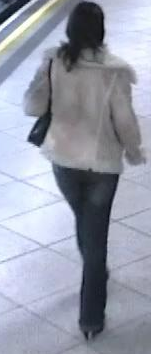}}
 \hspace{0.3cm}
\subfloat[]{\includegraphics[width= 1.6cm,height=3cm]{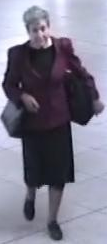}}
\hspace{0.3cm}
\subfloat[]{\includegraphics[width= 1.6cm,height=3cm]{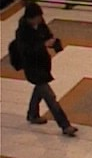}}
\hspace{0.3cm}
\subfloat[]{\includegraphics[width= 1.6cm,height=3cm]{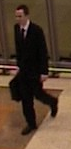}}\\
\vspace{-1.3\baselineskip}
\subfloat[]{\includegraphics[width= 1.6cm,height=3cm]{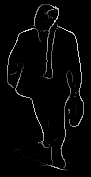}}
\hspace{0.3cm}
\subfloat[]{\includegraphics[width= 1.6cm,height=3cm]{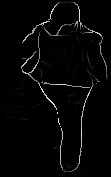}}
\hspace{0.3cm}
\subfloat[]{\includegraphics[width= 1.6cm,height=3cm]{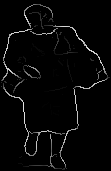}}
\hspace{0.3cm}
\subfloat[]{\includegraphics[width= 1.6cm,height=3cm]{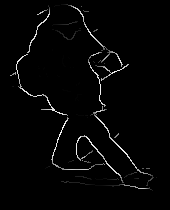}}
\hspace{0.3cm}
\subfloat[]{\includegraphics[width= 1.6cm,height=3cm]{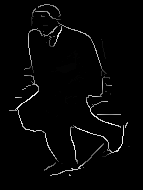}}\\
\vspace{-1.3\baselineskip}
\subfloat[]{\includegraphics[width= 1.6cm,height=3cm]{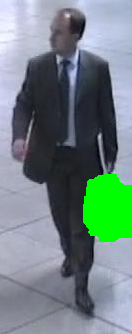}}
\hspace{0.3cm}
\subfloat[]{\includegraphics[width= 1.6cm,height=3cm]{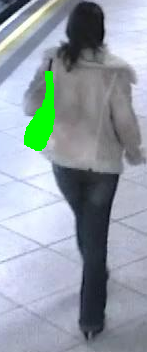}}
\hspace{0.3cm}
\subfloat[]{\includegraphics[width= 1.6cm,height=3cm]{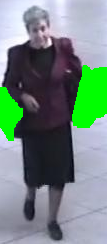}}\hspace{0.3cm}
\subfloat[]{\includegraphics[width= 1.6cm,height=3cm]{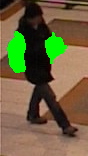}}
\hspace{0.3cm}
\subfloat[]{\includegraphics[width= 1.6cm,height=3cm]{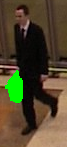}}\\
\vspace{-1.3\baselineskip}
\subfloat[]{\includegraphics[width= 1.6cm,height=3cm]{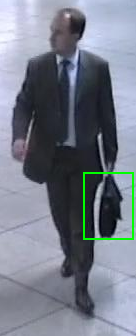}}
\hspace{0.3cm}
\subfloat[]{\includegraphics[width= 1.6cm,height=3cm]{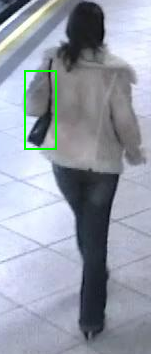}}
 \hspace{0.3cm}
\subfloat[]{\includegraphics[width= 1.6cm,height=3cm]{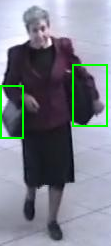}}
\hspace{0.3cm}
\subfloat[]{\includegraphics[width= 1.6cm,height=3cm]{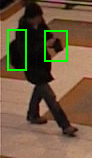}}
\hspace{0.3cm}
\subfloat[]{\includegraphics[width= 1.6cm,height=3cm]{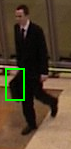}}
\caption{Some examples of our results on five frames of PETS 2006 and i-Lids AVSS videos. (Top row) original images. (Second row) extracted edge maps using the method of \cite{Belongie}. (Third row) segmented COs and (Last Row) detected COs by our method.}
\label{A2:ExampleResult}
\end{figure}

We are interested in identifying objects carried by people without any assumption on what they are. The top two rows of Figure ~\ref{A2:ExampleResult}  show people carrying different objects with their extracted edge maps, while the bottom two rows show the segmentation results and bounding rectangle detected by our method. Figure ~\ref{A2:ExampleResult}  also illustrates some of the challenges of carried object detection such as missing edges along the carried object outline, various carried object shapes and occlusion. To make our method applicable to more scenarios, we designed a method that can detect carried objects from a single frame without the use of temporal information.

Detecting carried objects (COs) is a challenging problem due to the large number of possible items people can carry, the variations in the way they can be carried and occlusion. To make this problem easier, a number of researchers assume that the objects are relatively large or that they are significantly protruding from a person's body. Assuming that possible carried objects are limited to a few classes, some researchers \cite{Branca, Dondera} rely on pre-trained object models. In these approaches, carried object models are learned based on a pre-existing set of known object models. However, because there can be a large variety of carried objects for which the type is usually unknown in advance, pre-learned models for carried objects have limited applicability. A group of methods tackle this problem using instead prior knowledge of the human body shape \cite{Damen, Haritaoglu, Chayanurak}. Relying on the hypothesis that a human body outline is symmetrical with respect to the body axis and that the limbs have periodic motion, aperiodic and asymmetric parts are considered as potential regions for carried objects \cite{Haritaoglu, Chayanurak}. Indeed, a backpack will be visible on one's back when the person is viewed from the side. However, COs do not always alter the human silhouette depending on the person's position with respect to the camera or the spatial relation of COs with the person's body. In addition, any deformations of human body outline may not always be a sign of a carried object. Person's clothes can also alter the human shape.

In our previous work (ECE) \cite{Ghadiri2016}, we introduced a carried object detector based on an ensemble of contour exemplars. In contrast with many other carried object detectors that rely on videos, this method is able to detect carried objects without temporal information, which makes the method applicable to still images. In ECE, we modeled usual human contours and compared them to a set of contours of a target person for CO detection. Contours that do not fit with the model are further analyzed to detect carried objects. Despite the success of our previous method to cope with many problems of carried object detection such as limitations caused by assumptions on carried object size, location and protrusion, improvements are needed to cope with some problems that affect the generated human model. In this work, we extend our ECE method by now identifying which superpixel belongs to either a person or a carried object. A superpixel provides a reference point and scale for the shape context descriptor to extract more meaningful information of contours which leads to large improvement in human shape modeling as we will show in section~\ref{sec:resanddic}.

We aim to detect any carried object without a specific model of what to look for. Our main idea is to leverage both top-down and bottom-up cues to detect human-like regions, and then detect COs using those regions are less likely to be a human region. To implement this idea, we first model different walking/standing poses using a codebook of local features and their spatial relationship to the centroid of the person. The local features are formed by extracting a Shape Context (SC) feature for each superpixel of an image region over-segmented at various scales. During COs detection, we match each extracted feature from each superpixel in the test image to the codebook. The matching quality reflects the likelihood of a superpixel to belong to a human region. We then use the complement of this likelihood to detect COs. We generate a likelihood map for carried objects by first combining the COs superpixel-wise probability maps of all the scales into a pixel-level probability map. Highly probable superpixels in the carried object likelihood map for which the boundaries are well aligned with the edge map are grouped to form carried objects.

To the best of our knowledge, no prior work explores human body modeling by multi-scale superpixel shape information to differentiate between a person region and carried object region. Towards this goal, the novel contributions of our technique are: (1) a novel multi-scale superpixel feature extraction method, and (2) detection of carried objects without any assumption on their locations and their shapes. We applied our detector to video frames, analyzed its components in detail, and show state-of-the-art results.

The rest of the paper is structured as follows. Section \ref{A2:relwork} reviews related work, section \ref{A2:proposed} describes the motivation of superpixel-based detection and the details of the superpixel labeling method to build a human model and to detect carried objects, section \ref{A2:experiment} justifies the contribution of the proposed method by qualitatively and quantitatively comparing its performance with the performance of several other state-of-the-art methods on two datasets, and finally in Section \ref{A2:conclude}, the paper is concluded.

\section{Related Work}
\label{A2:relwork}
There has been a growing interest in detecting carried objects from video frames. One of the first approaches on the subject was Backpack by Haritaoglu et al. ~\cite{Haritaoglu}. Backpack detects COs by symmetry and motion analysis of a person body. This work is built upon the assumptions that the human body shape is roughly symmetrical around a vertical axis, and limbs exhibit periodic motion when walking without objects. A dynamic model called temporal template is constructed by aligning and averaging foreground blobs of a person in short video sequences. Next, non-symmetric regions of the temporal template are segmented by calculating a vertical symmetry axis using Principal Component Analysis (PCA) and analyzing the distance of the pixels of the silhouette to the body axis. Periodicity analysis is applied on the non-symmetric regions to distinguish between regions that belong to limbs and carried objects. Javed et al.~\cite{JavedOmar} pointed out that calculating major axis using PCA is not a good way to obtain the vertical symmetry axis because the shape of the silhouette is deformed and hence the symmetry axis is displaced if a person carries an object. They suggest that the vertical line passing through the centroid of the head is a better candidate for the symmetry axis. To estimate repetitive motion of a person body, a feature vector called Recurrent Motion Image (RMI) is introduced. Non-symmetric regions that show periodic motion in the RMI are detected as limbs and the rest are considered as COs.

As the silhouette shape is well-descriptive of the human object class, several research directions have been explored about using prior information of human body shape to distinguish it inside a foreground blob and to explain the remaining foreground parts in terms of carried objects. Lee et al.~\cite{Lee2006} detected carried objects as outliers in the extracted foreground using a dynamic shape model of human motion. Dynamic shape deformations of different people in different viewpoints are modelled using kinematic manifold embedding and kernel mapping. Carried objects are mismatching parts in the person silhouette when compared with the best matching dynamic shape model. In the work of Chayanurak et al.~\cite{Chayanurak}, the human silhouette is represented by a star skeleton. The authors detected the people carrying objects by analyzing a time series of motions of extracted skeleton limbs. Those tracked limbs which are motionless or moving with the overall human body are detected as carried objects.

The work of Damen and Hogg~\cite{Damen} uses several exemplar temporal templates of unencumbered people in different viewpoints to compare with built temporal template of a person in the test set. Given a tracked person, a temporal template is created and viewing direction of the person is computed using homography information. Next, for detection, an exemplar with the same viewpoint as the tracked person is chosen and scaled to best fit the temporal template. Carried object candidates are mismatching parts in a tracked temporal template compared with the best matching exemplars. They are regions that do not exhibit significant periodicity and fit in the spatial prior map. The idea was further developed by Tzanidou et al.~\cite{Tzanidou}. The authors assumed that clothing colors differ from COs colors, so they use a color temporal template to exploit color information to better discriminate the human body from COs. They also automated the exemplar matching step of Damen and Hogg~\cite{Damen} by computing viewpoints using shoulder shape features and machine learning algorithms. Unfortunately, the performance of these last two systems is rather poor when a prior spatial model for carried object is not provided. In the work of Tavanai et al.~\cite{Tavanai} a carried object is characterized in terms of convex and elongated shape models instead of pre-trained models. Under the assumption that carried objects are usually protruding from a person body and that they follow the person trajectory with continuous and spatially consistent characteristics, convex or elongated objects that are not carried objects can be rejected.

Aforementioned methods rely on a precise foreground segmentation which is often difficult to achieve in video surveillance due to sudden change of lighting or poor contrast from the surrounding environment. To overcome this problem, Senst et al. ~\cite{Senst2012} described a pedestrian by motion statistics using Gaussian mixture motion model and calculating short-term and long-term uniformity of motion. Short-term uniformity of motion models the periodic motion of limbs, and long-term uniformity of motion models uniform motion and corresponds to head and torso. If a person carries an object, her uniform motion profile does not fit to the average motion profile of unencumbered people. Those parts of a person that do not fit with both models are considered as carried objects.

Several approaches have been focusing on detecting specific classes of carried objects. In Branca et al.~\cite{Branca} work, two classifiers are trained to detect two types of tools that are often used by intruders on archaeological sites. They use the wavelet transform coefficients of a binary foreground blob as features to feed the classifier and search around a person body to detect COs. Chua et al.~\cite{Chua} detected sling bag and backpack using their geometrical shape cues (e.g., sling bag straps can be described as two near-parallel lines). Yue et al.~\cite{Yue} modeled the spatial relation of points on the person contour to the person main axis using Support Vector Machine (SVM) to classify backpacks and luggage. There are some works on semantic segmentation (\cite{DBLP,Parsing,SegNet}) that are using baseline systems, such as the Fast/Faster R-CNN \cite{FRCNN} and Fully Convolutional Network (FCN) \cite{FCN} in which specific classes of carried object are segmented. Liang et al.~\cite{Yamaguchi}  used a contextualized convolutional neural network to decompose a human image into semantic clothes/body regions such as face, dress, bag, etc. They applied their network on the Fashionista dataset~\cite{Fashionista} in which the images are near frontal-view and collected from Chictopia.com, a social networking website for fashion bloggers.  Kaiming He et al.~\cite{DBLP} extended Faster R-CNN \cite{FRCNN} for object instance segmentation by adding a branch for predicting segmentation masks. They evaluated their work on COCO dataset~\cite{Lin2014} which contains 91 object categories such as person, bicycle, cellphone, bags, etc.

Some studies tried to detect people carrying objects without specifically localizing them. Abdelkader and Davis~\cite{Abdelkader} built a model of a natural gait by dividing the human body into four horizontal segments and measuring the widths of the segments over several frames. They formulated two constraints on the amplitude and periodicity of this model and assumed that violation of these constraints corresponded to a person carrying an object. Delgado et al.~\cite{Delgado} first segmented images into salient regions by clustering superpixels using~\cite{Sander} under the assumption that the background, carried objects and clothes have discriminative colors. These regions are used to generate a contour map by extracting the boundaries of the obtained regions. Tiling the detection window with a grid of Histogram of Oriented Gradients (HOG) descriptors and using concatenated feature vectors in a SVM based window classifier allowed them to detect three classes of COs: low lateral bag, shoulder bag, backpack. A recent work by Wahyono ~\cite{Wahyono2017} classified human carrying baggage using convolutional neural network (CNN). In this work, a simple convolutional neural network architecture consisting of two convolutional layers followed by max-pooling layers is proposed. A detected gray-scale moving person is divided into three parts of top, middle and bottom sub-images and then combined to form a three channel image to be fed into the network. Their experimental results show that their CNN feature based classifier outperforms the best achievement by conventional methods to classify a person's carrying baggage. However, the object is not localized.

Approaches based on the shape of a person silhouette fail to detect COs with little or no protrusion. Dondera et al.~\cite{Dondera} suggested that segmenting a silhouette based on color and motion information and then analyzing each region could be a solution. They segmented a silhouette into regions using three detectors: a segmentation-based color detector, an occlusion boundary-based moving blob detector, and an optical flow-based protrusion detector. Then, they fed a Gaussian kernel SVM with the extracted features (e.g., compactness, distance of the region to the center of person) from each region to classify them as carried object or not. However, because of large variations in COs appearance and similarity to other items of clothing, the effectiveness of the features extracted from each region is questionable.

Ghadiri et al.~\cite{Ghadiri2016} detected carried objects by analyzing a person contours instead of its silhouette. The authors implemented their method without any protrusion assumption and also without prior information about the COs. Exploiting contours instead of silhouette-based characteristics such as symmetry or shape, enabled the proposed method to detect non-protruding objects contrarily to most state-of-the-art methods. A decision about carried objects is made based on only one frame of the entire video. This work was extended in~\cite{Ghadiri2017} by integrating static and dynamic cues  to use temporal information. Their goal was to detect and segment carried objects using a short video sequence of a person. To this end, they first detected several candidate regions for carried objects in each frame of a person's short video sequence, then using spatio-temporal information of the candidate regions, consistency of recurring carried object candidates viewed over time is obtained and serves to detect carried objects. Results were improved by using temporal information. However, this method cannot be applied to still images, so it is less general.

\section{Proposed method}
\label{A2:proposed}
We formulate carried object detection as a superpixel classification problem. Our goal is to find a subset of superpixels that belongs to a person or to the background, and explain the remaining superpixels in terms of carried objects.
There are three main steps in our approach: (1) Building a codebook of local features based on contour information of upright persons, standing or walking, without carried objects from different viewpoints; (2) Finding carried object-like superpixels using the codebook; (3) Applying contour assisted segmentation and detection. The final output is a pixel-level segmented carried object delineated with a bounding box. We will now describe each step.

\subsection{Building a codebook of local features}
\label{Sec:BC}
We model walking/standing people from different viewpoints by constructing a codebook of local features along with their positions with respect to the centroid of a person. In a manner similar to ECE \cite{Ghadiri2016}, we require the following data to learn the model:

\begin{itemize}
  \item Images of people delineated by bounding boxes.
  \item The associated segmentation masks of people that do not include COs.
\end{itemize}
 
We have chosen people exemplars from three different datasets: INRIA, PETS 2006 and i-Lids AVSS as training images and separated them into 8 viewpoints. A person in each image is described by a bounding box provided by \cite{DPBM}. Then the person is rescaled to a fixed size (in our case $170\times80$). Foreground masks\footnote{We use the term foreground mask to denote segmented areas containing a person's regions and possible carried objects} for images from INRIA dataset are obtained manually, while for PETS 2006 and i-Lids AVSS, they are obtained automatically by applying the PAWCS segmentation method \cite{Charles}. A still image foreground segmentation algorithm could also be used. For the selected exemplars, carried object regions are eliminated from the foreground masks because we want to build a model of a person without carried objects.

\subsubsection{Superpixel generation and feature extraction}
\label{Sec:SPG}
 
An input image (resized image of a person delineated by a bounding box), both for training and detection, is segmented into superpixels at multiple scales using Simple Linear Iterative Clustering (SLIC) \cite{Achanta} as implemented in the VLFeat toolbox. Superpixel sizes across the scales are chosen to ensure that object boundaries are reasonably well approximated by superpixel boundaries. Then, boundary edges in the foreground mask are extracted by applying \cite{Martin}. Each superpixel ($SP_i$) is then characterized with a Shape Context (SC) feature by Wang et al. \cite{Wang} extracted at its center. The intuition behind our feature extractor is that all pixels in a superpixel belong to the same object which leads to the advantage of gathering shape information on a perceptually meaningful atomic region. 

\begin{figure}[hht]
\captionsetup[subfigure]{labelformat=empty}
\centering
{\includegraphics[width= 10.5cm,height=5cm]{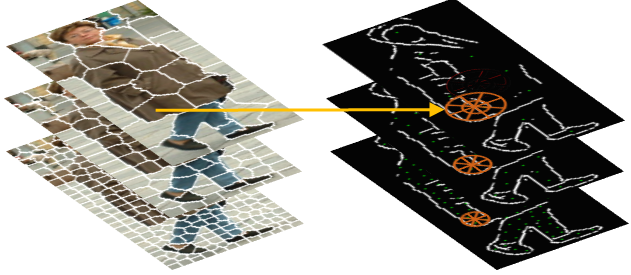}}
\caption{An example of generated superpixels at three different scales. The three levels of coarse-to-fine superpixels of a person are shown at the left (The highest layer is the coarsest level $L_3$). On the right, an extracted edge map is shown with the shape context descriptor corresponding to a specific superpixel.}
\label{A2:SPlayer}
\end{figure}

In our experiments, three levels of coarse-to-fine superpixels ${L_3,L_2,L_1}$ are generated, and for each level, the SC size is chosen based on the superpixel size. To compute SC feature at the center of a superpixel, only edges that belong to the dilated superpixel region are counted. Dilation of a superpixel compensates for small misalignment between objects and superpixels boundaries. Figure~\ref{A2:SPlayer} shows an example of generated superpixels for a person in PETS 2006 dataset at three scales and a shape context descriptor for a superpixel.

\subsubsection{Offline learning}
\label{Sec:learning}
In the offline training phase, each SC feature vector extracted from a training image with its position relative to the person center is stored in the codebook along with the viewpoint of the person. Therefore, each codeword $ce_{j}=({sc_j^{ce}, d_j^{ce}, v_j^{ce}},L_j^{ce})$ records four types of information of superpixel $SP_j$: shape context feature $sc_j^{ce}$, relative distance $d_j^{ce}$ of the center of the superpixel $SP_j$ to the centroid of the person bounding box, viewpoint of the person $v_j^{ce}$, and $L_j^{ce}\in({L_1,L_2,L_3})$ the superpixel scale. Viewpoints are quantified into 8 directions. A codeword for a superpixel that carries little information because of a small person boundary support (few edges are falling in the SC bins) is removed from the codebook.

Each person exemplar in the training set has possibly a unique clothing style and a unique standing or walking pose. Therefore, most extracted codewords are representative of a specific exemplar. In other words, due to the small number of exemplars available from different viewpoints, applying a clustering approach to the learned representative codewords would lead to missing shape information. As a results, we keep almost all SC features $(sc_j^{ce})$ to include many possible deformations except the less informative (sum of shape context feature is smaller than a fixed threshhold) SC features. In addition, the contour position relative to the center of a person plays an important role in finding a matching codeword from the codebook. Therefore, bag of words approaches are not applicable for our purpose.

\subsection{Finding carried object-like superpixels}
\label{FindMatch}
Given a test image, a person is detected by applying a window-based detector \cite{DPBM}. We obtain the foreground mask using PAWCS \cite{Charles}. Since, a carried object may protrude from the obtained bounding box, the foreground mask that overlaps best with the person bounding box is selected as a region of interest (ROI). We assume that the detected person is not occluded. The ROI contains a person, potentially one or more carried objects and parts of the background region because segmentation methods do not output perfect foreground masks. Using an estimation of the person scale provided by the person detector, the ROI is scaled to a predefined size so that the person fits inside a $80\times 170$ window. Then, we segment the ROI into superpixels at different scales.
Given an ROI represented by superpixels, our goal is to assign a label (person or background or carried object) to each of them. Each superpixel is scored based on the matching of its SC feature with the codebook to determine how likely it is to belong to a person. To this end, each extracted $sc_i^{R}$ from the center of superpixel $SP_i$ in the ROI is only compared to the codebook entries for which the distance difference of their relative position to the center of the person falls in a small ellipsoid area with minor axis=6 and major axis=15. Therefore, the matching score for a superpixel $SP_i^{R}$ in a specific superpixel scale level $L_i$ is computed as follow:
\begin{equation} \label{A2:matchce1}
P(SP_i^{R}|d_i^ {R},L_i)={exp(-{\lVert sc_j^ {ce}-s_i^ {R}\lVert}) P(d_i^ {R}|d_j^ {ce},  \Sigma)}P(L_i)
\end{equation}

with
\begin{equation} 
P(d_i^ {R}|d_j^ {ce}, \Sigma)=\dfrac{1}{2\pi\sqrt{| \Sigma|}}exp(-\dfrac{1}{2}{(d_i^ {R}-d_j^ {ce})}^T \Sigma^{-1}(d_i^ {R}-d_j^ {ce}))
\label{A2:matchce2}
\end{equation}

where $d_i^ {R}$ ($d_j^ {ce}$) is a relative distance of the $sp_i$ ($sp_j$) to the center of a person bounding box in the test image (codeword). $\Sigma$ is a $2\times2$ diagonal covariance matrix with ${ \Sigma_{11}}<{ \Sigma_{22}}$. Because the human body is fitted better by an ellipsoid, an ellipsoid area is chosen over circular area to compute $P(d_i^ {R}|d_j^ {ce})$. The $k$ best matching codewords for a $SP_i^{R}$ are selected and stored in set $S_{i\in(1...N)}^{R}=({sc_1^{ce},sc_2^{ce},...,sc_k^{ce}})$ based on Equation ~\ref{A2:matchce1}, where $N$ is the sum of superpixels at three different levels and $k=80$ (see section~\ref{Art2:smoothing}). Further analysis is applied on $S_{i\in(1...N)}^{R}$ to estimate the superpixel probability. This process is applied independently to each superpixel scale.

\subsubsection{Person viewpoint}
\label{Sec:vp}
Contours of anatomical parts, such as arms and legs as well as some clothing parts may look like carried object contours in a side view of a person if they are compared to a frontal view model. To alleviate this problem, we only consider the matching contours that both come from the same viewpoint. For this reason, the viewpoint of a person in the test set is estimated by computing the most frequent person's viewpoint in the obtained set $S_{i\in(1...N)}^{R}$ of the previous step. Once the viewpoint is calculated, the probability value for a superpixel $SP_i^{R}$ with a viewpoint $v_p$ is computed by the maximum probability of set $V$, where $V$ is a subset of set $S$ ($V\subset S$) where in set $V$, person's viewpoint is similar to the obtained viewpoint $v_p$.

\begin{equation} 
P(SP_i^{R}|v_p,L_i)=\max_{j} {exp(-{\lVert sc_j^ {V}-s_i^ {R}\lVert}) P(d_i^ {R}|d_j^ {V},  \Sigma)P(L_i)}
\label{A2:Probanility}
\end{equation}

\subsubsection{Carried object-like superpixels} \label{Art2:smoothing}
\begin{figure}[htbp]
\captionsetup[subfigure]{labelformat=empty}
\centering
{\includegraphics[width= 10cm,height=6.3cm]{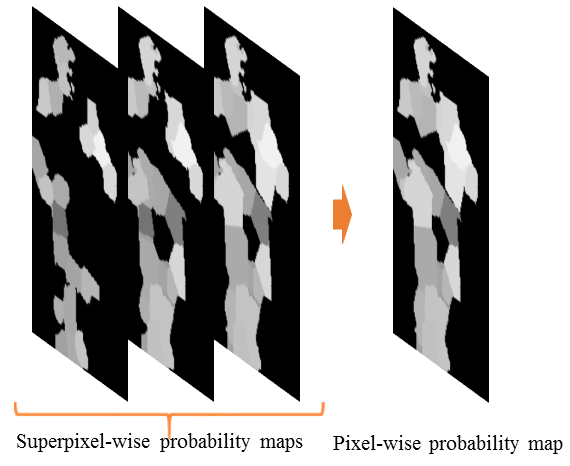}}
\caption{An example of generated pixel-wise probability map by maximizing the superpixel-wise probability over three superpixel scales.}
\label{A2:fusing}
\end{figure}

At this stage, we have a probability map for each superpixel scale level. A probability map indicates the probably that a superpixel correspond to a part of a person. The three levels of the superpixel-wise probability maps are fused to generate a pixel level probability map by maximizing the superpixel-wise probability over the three level ($max_{k}(P(SP_i^{R}|v_p,L_k))$), where $k\in({1,2,3})$ (see Figure~\ref{A2:fusing}).

\begin{figure}[htbp]
\captionsetup[subfigure]{labelformat=empty}
\centering
{\includegraphics[width= 9.5cm,height=6.5cm]{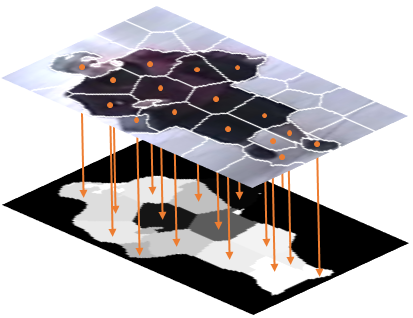}}
\caption{An example of generated superpixel-wise probability map by smoothing the probability of each superpixel in the coarsest superpixel scale.}
\label{A2:probLayer}
\end{figure}

We use the pixel-level probability map and superpixel at the coarsest level ($L_3$) to generate a final probability map. To this end, we use superpixels in the third level $L_3$ to smooth the obtained pixel-wise probability map by assigning the same probability to each pixel belonging to the same superpixel. This probability is computed by averaging the probability of all pixels in the superpixel (Figure~\ref{A2:probLayer}).

\begin{figure}[htbp]
\captionsetup[subfigure]{labelformat=empty}
\centering
\subfloat[a]{\includegraphics[width= 3cm,height=4cm]{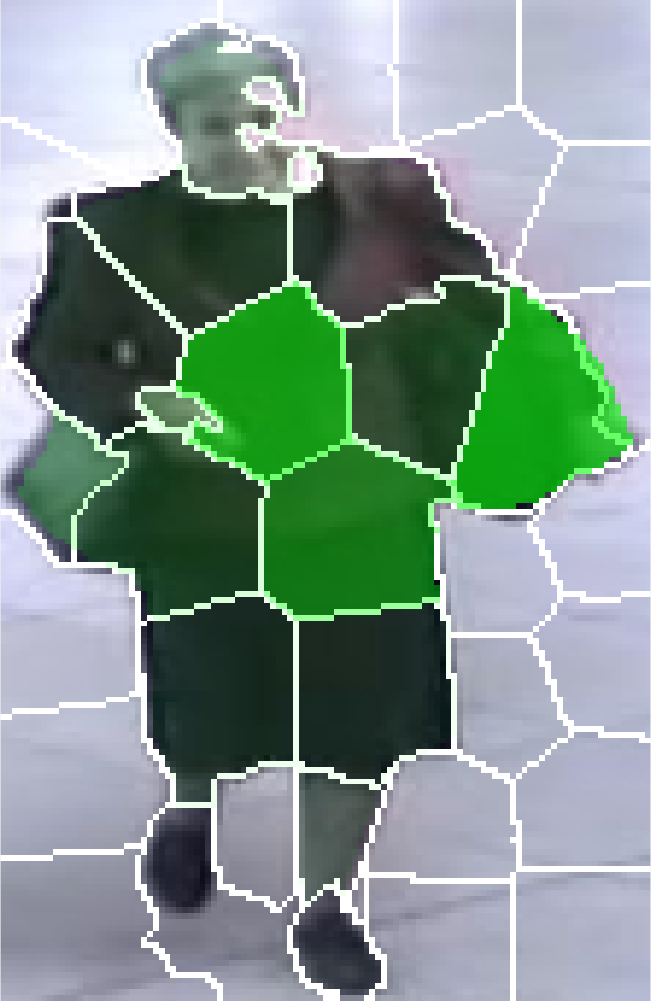}}
\hspace{0.3cm}
\subfloat[b]{\includegraphics[width= 3cm,height=4cm]{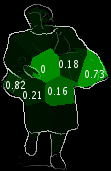}}
 \hspace{0.3cm}
\subfloat[c]{\includegraphics[width= 3cm,height=4cm]{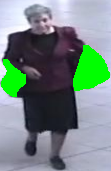}}
\hspace{0.3cm}
\caption{Carried object probability maps. Lighter green mean higher CO probability (a) overlaying of the carried object probability map on original image, (b) overlaying of the CO probability map on the edge map with boundary support values, (c) detected carried objects.}
\label{A2:COoverlay}
\end{figure}

As we are doing binary classification (body parts or carried objects in the ROI), the complement of the person probability map is considered as the carried object probability map. Figure~\ref{A2:COoverlay} shows carried object probability overlaid on a test image.

At this step, we also make use of a similar technique as mentioned above to refine the foreground mask. As mentioned earlier, the foreground mask may also include some parts of the background because of foreground segmentation errors. The procedure described above to generate carried object probability map only considers two classes (person, carried object), so misclassified background regions may also get high probability. To remove these misclassified region, superpixels in the third level ($L_3$) are overlaid on the foreground mask. Then, for each superpixel, the proportion of foreground pixels over the superpixel area is computed. If the obtained proportion is less than $0.75$, the superpixel is assigned to background. In other words, each superpixel should only belong to the background (bg=0) or foreground class (fg=1). 
 
\subsection{Contour-Assisted Segmentation}
\label{A2:Contour_assist}
At this stage, we have a superpixel-wise probability map of carried objects. To obtain the final carried object segmentations, we attempt to group superpixels. We first compute a superpixel boundary support based on the boundary edge map. The superpixel boundary support value is obtained by calculating the number of overlaps between the boundary pixels of a superpixel and edges in the edge map, divided by the perimeter of the superpixel. The main idea behind this is that a superpixel with a high boundary support is probably a CO or part of it. Figure~\ref{A2:COoverlay} shows the overlay of the carried object probability map on the edge map along with the boundary support value of the superpixels with the highest probabilities.

\begin{figure}[htbp]
\captionsetup[subfigure]{labelformat=empty}
\centering
\subfloat[a]{\includegraphics[width= 3cm,height=4cm]{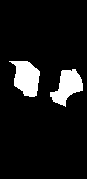}}
\hspace{0.3cm}
\subfloat[b]{\includegraphics[width= 3cm,height=4cm]{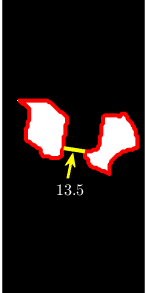}}
\vspace{-1em}
\\
\subfloat[c]{\includegraphics[width= 3cm,height=4cm]{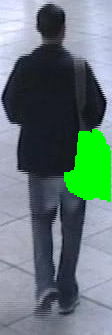}}
 \hspace{0.3cm}
\subfloat[d]{\includegraphics[width= 3cm,height=4cm]{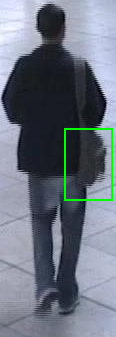}}

\caption{Final segmentation of carried object with non-maximum suppression. (a) initial segmentation of carried objects, (b) minimum distance between two segments (shown by yellow line), (c) segmented carried object after non-maximum suppression and morphological operations, (d) carried object bounding box.}
\label{A2:NMS}
\end{figure}

The initial segmentation is obtained by grouping neighboring superpixels with the highest probability values and the highest superpixel boundary support values. The superpixels with the highest probability values are found using threshold $T_p$ and the superpixels with the highest boundary support are found with threshold $T_c$, both determined experimentally. Figure~\ref{A2:NMS} (a) shows the initial segmentation of the carried objects. Note that we do not take into account any information of visual appearance, such as color or texture while grouping the superpixels. This is because usually a carried object does not correspond to a compact region with distinct grey level or color.

Using the initial segmentation, a final segmentation is obtained by applying a non-maximum suppression method. At this step, we keep only the segmented regions with a boundary support values that is larger than that of its neighboring regions inside a distance $D=15$. Figure~\ref{A2:NMS} shows two highly probable CO regions with their closest path to each other (in this case $D=13.15$). Detected regions by NMS method are then refined by applying a dilation operation by a disk of radius $3$ (Figure~\ref{A2:NMS} (c)).

\section{Experiments}
\label{A2:experiment}
In the experiments we compared our work, which we refer to as SuperPixel-based Carried Object Detector (SPCOD), to CO detection methods that, like our method, do detection without temporal information. Therefore, we compared our method with ECE \cite{Ghadiri2016} and a deep learning-based semantic segmentation method~\cite{SegNet} re-trained for CO detection (we will refer to it as SegNetCO). 

To better understand the impact of our feature extraction method on the performance, we also compared our work with a variant which is a combination of our feature extraction method to generate a person's hypothesis and the contour analysis of \cite{Ghadiri2016} to detect carried objects, henceforth referred to as SPECE (SuperPixel-based Ensemble of Contour Exemplars).

The performance of our algorithm is evaluated qualitatively and quantitatively on two datasets: PETS 2006 and i-Lids AVSS. We prepared a training set from three different datasets: INRIA pedestrian \cite{Triggs}, PETS 2006 \footnote{http://www.cvg.reading.ac.uk/PETS2006/data.html}, and i-Lids AVSS \footnote{http://www.eecs.qmul.ac.uk/~andrea/avss2007\_d.html}. Training images include a wide range of different clothing. For each person viewpoint, approximately 40 images were selected for training.

\subsection{Datasets}\label{Art2:DataSet}
The PETS 2006 dataset is a collection of seven sets of videos, and each set contains four videos (about 10-minute length with a resolution of $720\times576$) taken with four stationary cameras to monitor a train station. Similarly to previous work, we used the seventh set of videos, the third camera view, and the 106 identified individuals for our evaluation. Seventy-five individuals are chosen for testing and the remaining individuals are selected for training. Each person can carry more than one object and the objects include suitcases, handbags, boxes, and musical instruments. Ground-truth bounding boxes for 83 carried objects are manually provided online by \cite{Damen}.

Evaluation was also carried out on the i-Lids AVSS dataset, which contains three videos (at $720\times576$ resolution) taken with stationary cameras at a train station. We followed the evaluation protocol defined in \cite{Ghadiri2016} and used their annotations \footnote{https://sites.google.com/site/cosdetector/home}. Fifty-nine individuals among 88 were selected for testing. Individuals that are not in the test set are used for training. Both datasets were originally created with a goal of detecting abandoned objects in a train station.

Similarly to \cite{Ghadiri2016}, the INRIA dataset is used only in the training to complement frames from PETS 2006 and i-Lids AVSS. 

\subsection{Experimental method}
\label{sec:setup}

We segmented the image into superpixels with different sizes using SLIC. The reason that we chose SLIC to generate superpixels is that it is a well performing superpixel segmentation algorithm that is relatively fast. The SLIC parameters are the region size and the regularizer. For our experiments, we have set region size $S_L$ proportional to the predefined size for a person ($80\times 170$), that is $S_{L1}=15$, $S_{L2}=20$, $S_{L3}=25$ for the three layers and the regularizer was set to 5000. Since objects rarely have wiggly shapes, choosing a high value for the regularizer generates superpixels with regular shapes that are less likely to cross object boundaries.

To capture the shape of contours along boundaries of superpixels, shape context is used which is a simple and robust descriptor to capture shape. More importantly, we can adjust shape deformation tolerance by the number and the size of SC bins. Size of the shape context for each superpixel layer is chosen slightly bigger than the superpixel region size to ensure that a superpixel shape can be captured by SC. Other parameters are $T_c=0.6$ and $T_p=0.74$. It should be noted that all parameters are the same for PETS 2006 and i-Lids AVSS.

In SPECE, the probability map for a person is generated using the technique explained in section \ref{FindMatch}. Contours that are not falling in the generated map are considered as candidates for carried object contours. Then, similar post processing as ECE is applied on the candidate contours using their publically available code. A region is assigned to each carried object candidate contour using Biased Normalized Cut (BNC) with a probability obtained by a weighting function of its overlap with the person's contour hypothesis map and segmented foreground. Finally, carried objects are detected by applying a Non-Maximum Suppression (NMS) method which eliminates the low score carried object candidates.

\begin{figure}[htbp]
\captionsetup[subfigure]{labelformat=empty}
\centering
\subfloat[]{\includegraphics[width= 2.3cm,height=3.5cm]{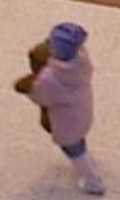}}
\hspace{0.3cm}
\subfloat[]{\includegraphics[width= 2.3cm,height=3.5cm]{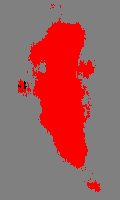}}
\hspace{0.3cm}
\subfloat[]{\includegraphics[width= 2.3cm,height=3.5cm]{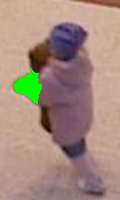}}
\vspace{-1.5\baselineskip}
\\
\subfloat[]{\includegraphics[width= 2.3cm,height=3.5cm]{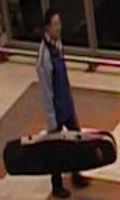}}
\hspace{0.3cm}
\subfloat[]{\includegraphics[width= 2.3cm,height=3.5cm]{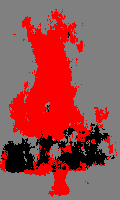}}
\hspace{0.3cm}
\subfloat[]{\includegraphics[width= 2.3cm,height=3.5cm]{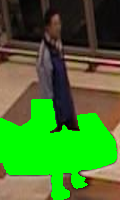}}
\vspace{-1.5\baselineskip}
\\
\subfloat[a]{\includegraphics[width= 2.3cm,height=3.5cm]{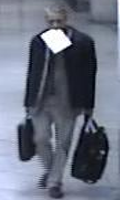}}
\hspace{0.3cm}
\subfloat[b]{\includegraphics[width= 2.3cm,height=3.5cm]{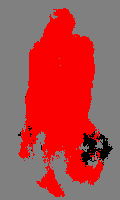}}
\hspace{0.3cm}
\subfloat[c]{\includegraphics[width= 2.3cm,height=3.5cm]{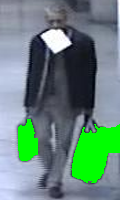}}

\caption{Detected carried object by SegNetCO. The first column shows three examples of a person in i-Lids AVSS and PETS 2006. The second column shows the pixel-wise segmentation of three classes of background, person and carried object with the grey, red and black colors respectively (SegNet output). Third column shows the segmentation results after the within-superpixel smoothing process.}
\label{A2:DeepPIC}
\end{figure}

In SegNetCO, we used SegNet architecture by~\cite{SegNet} and trained the network with the Fashionista dataset \cite{Fashionista}. Since there is no available dataset with sufficient size in the domain of carried object detection, we found Fashionista dataset useful to serve our purpose. Fashionista dataset includes mostly frontal view of upright people in fashion pictures. We modified the annotations provided in the Fashionista dataset so that all clothing items and the person's body parts are labeled as the single class of human, and different types of bags carried are classified as carried objects. After training, the network is fine tuned using our training set as presented in section \ref{Art2:DataSet} by keeping the earliest (first) layer of the network fixed (to avoid overfitting) and fine-tuning the rest of the layers. In the test phase, the detected moving object is resized to $200 \times 100$ and fed into the network. As it can be seen in Figure \ref{A2:DeepPIC}(b), CO regions is poorly segmented by SegNetCO. To connect the scattered pixel-level segmentation by SegNetCO, a dilation operation by a disk of radius $5$ is applied. Then, final segmentation results are further refined by applying the following within-superpixel smoothing method. We use superpixels in the third level $L_3$ to smooth the obtained pixel-level segmented image by assigning the same label to each pixel belong to the same superpixel (see Figure \ref{A2:DeepPIC}). Each superpixel is classified in the class of carried object if more than $20$ percent of its pixels are belonging to the carried object class.   

For a quantitative evaluation of the results, we use the F-measure, $F = \frac{2RP}{R+P}$ , where $R$ and $P$ are recall and precision, respectively. The overlap to determine true positive (TP) between a ground-truth CO and a detected CO is given by
\begin{equation} 
\frac{b_d\cap b_{gt}}{b_d\cup b_{gt}}>K
\label{A2:Eval}
\end{equation}

where $b_d$ and $b_{gt}$ are bounding boxes of the detected object and ground truth, respectively, and $K$ is a threshold value between zero and one. Then, if $T_{gt}$ is the number of ground truth COs, then $R=\frac{TP}{T_{gt}}$ and $P=\frac{TP}{TP+FP}$. It should be noted that additional detections of the same COs are counted as false positives.

\subsection{Results and discussion}
\label{sec:resanddic}

\begin{figure*}[htbp]
\captionsetup[subfigure]{labelformat=empty}
\centering
\subfloat[][(a)]{\includegraphics[width=0.48\linewidth]{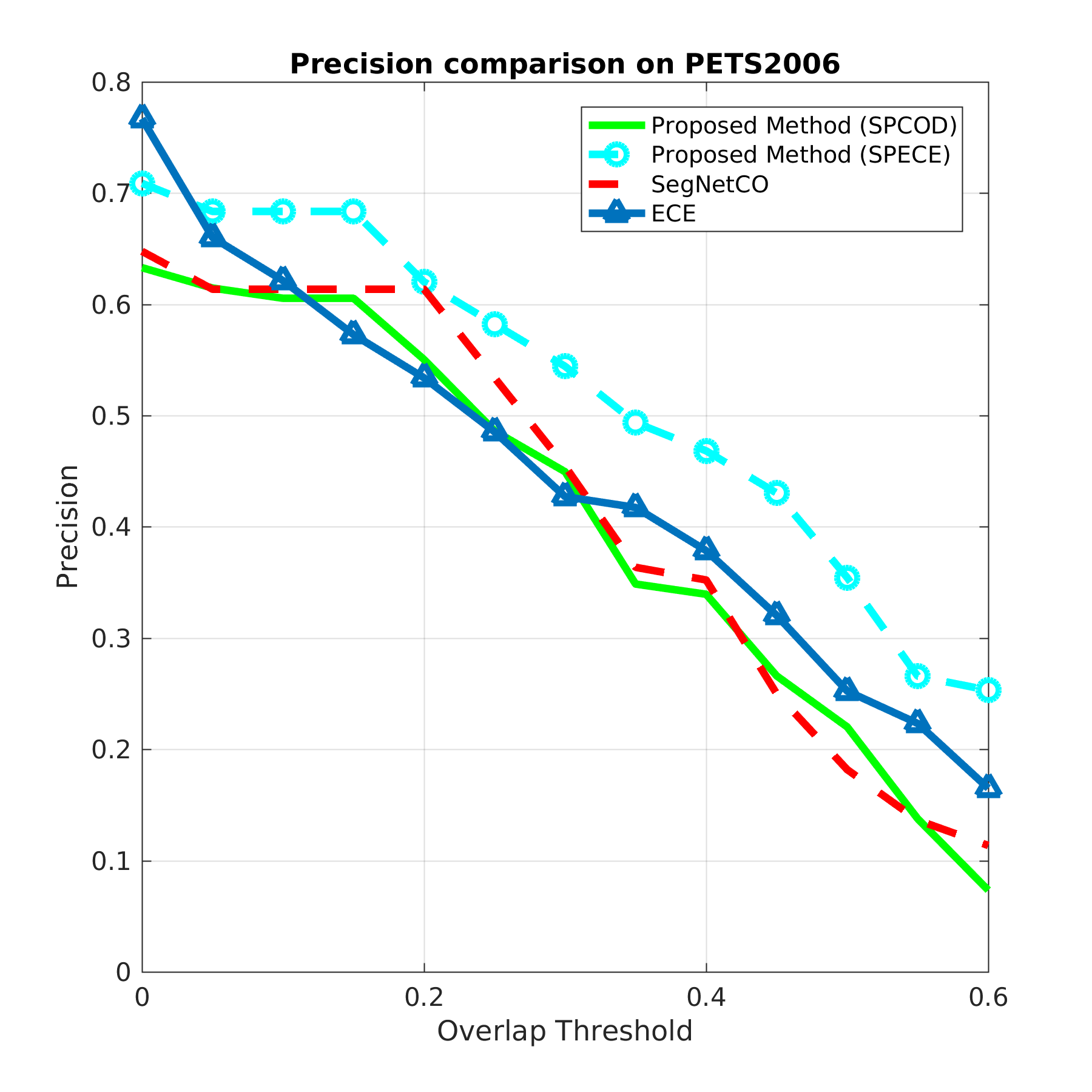}} \hspace{0.06cm}
\subfloat[][(b)]{\includegraphics[width=0.48\linewidth]{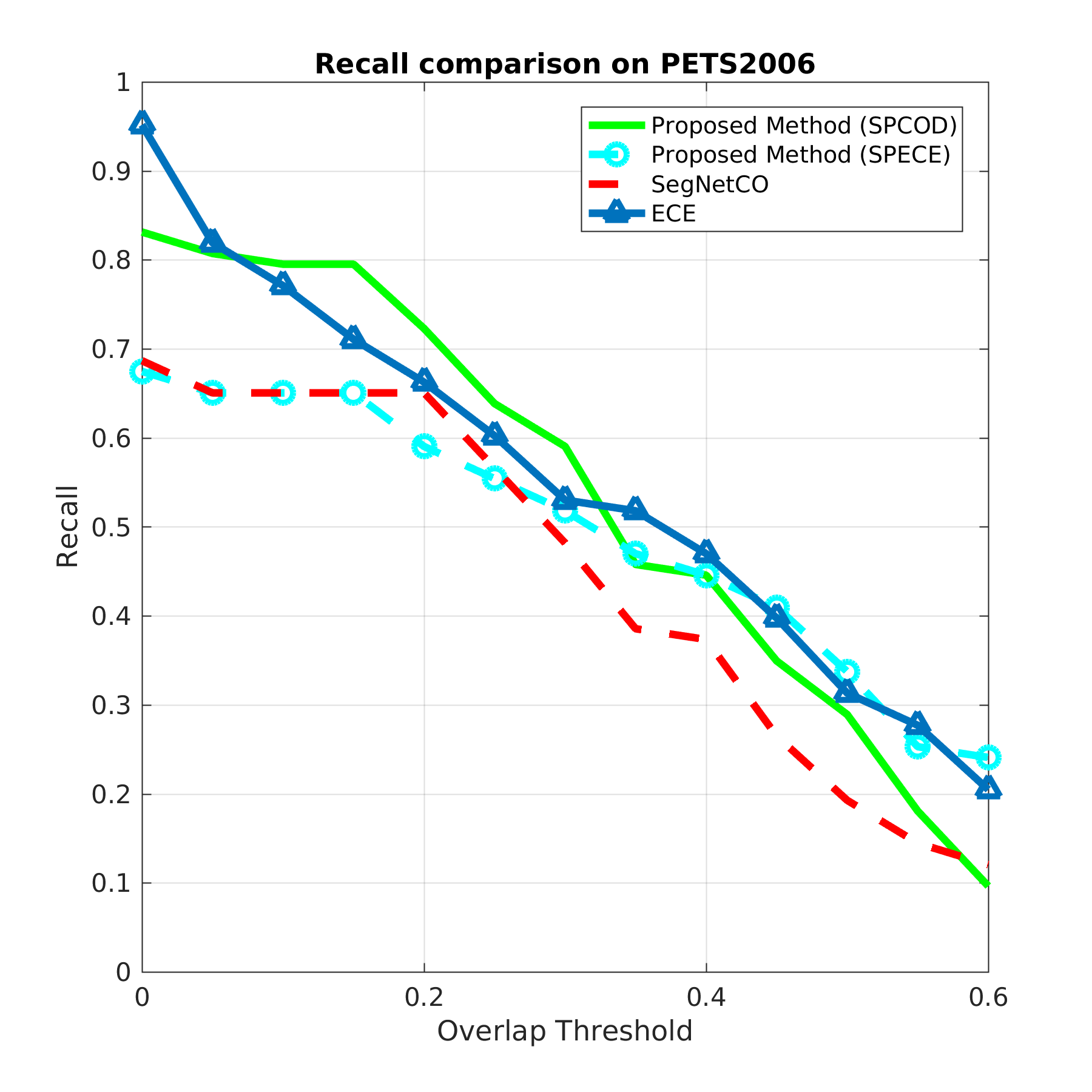}}

\caption{Precision and recall plots as function of the overlap threshold on PETS 2006.}
\label{A2:Per:P}
\end{figure*}

\begin{figure*}[htbp]
\captionsetup[subfigure]{labelformat=empty}
\centering
\subfloat[][(a)]{\includegraphics[width=0.48\linewidth]{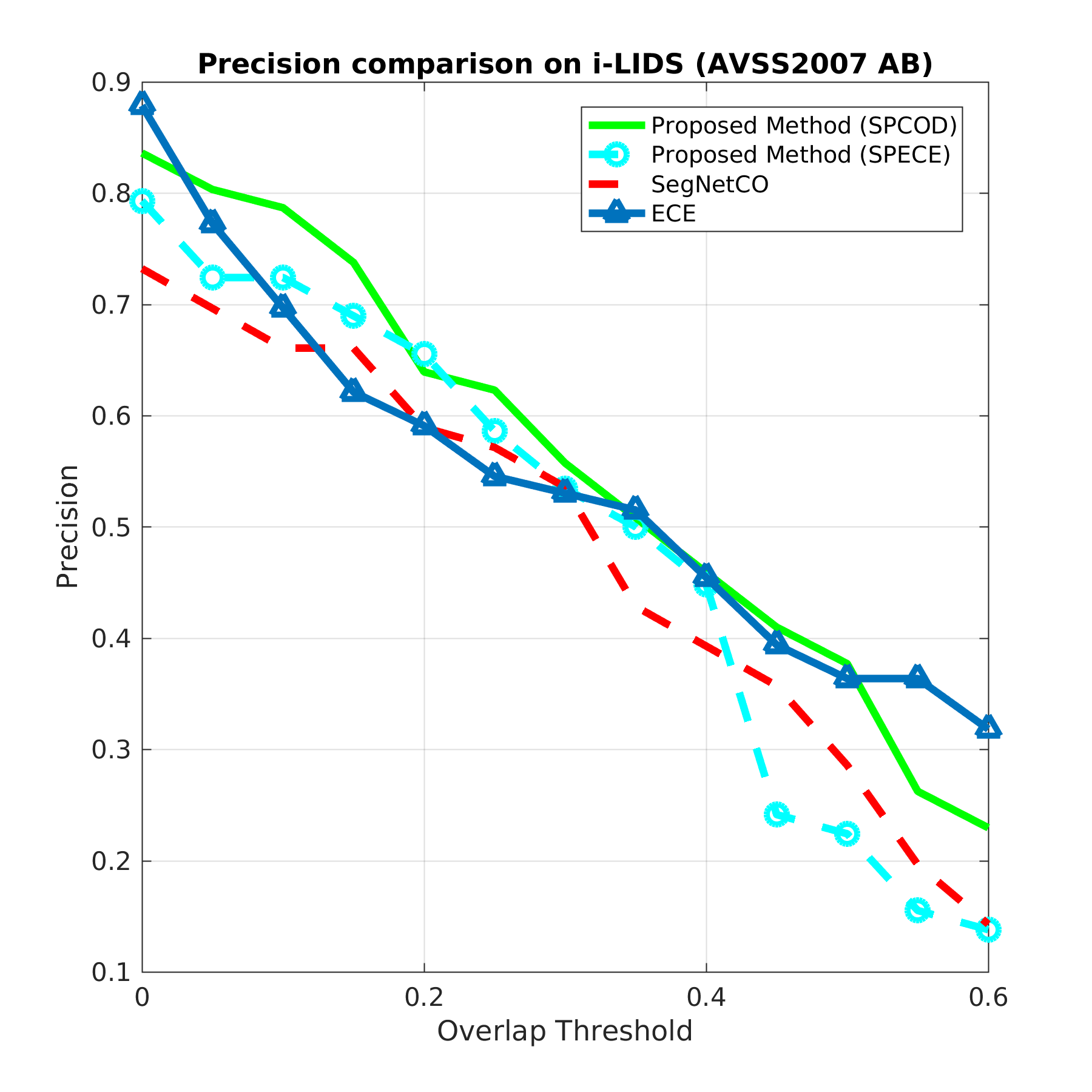}} \hspace{0.06cm}
\subfloat[][(b)]{\includegraphics[width=0.48\linewidth]{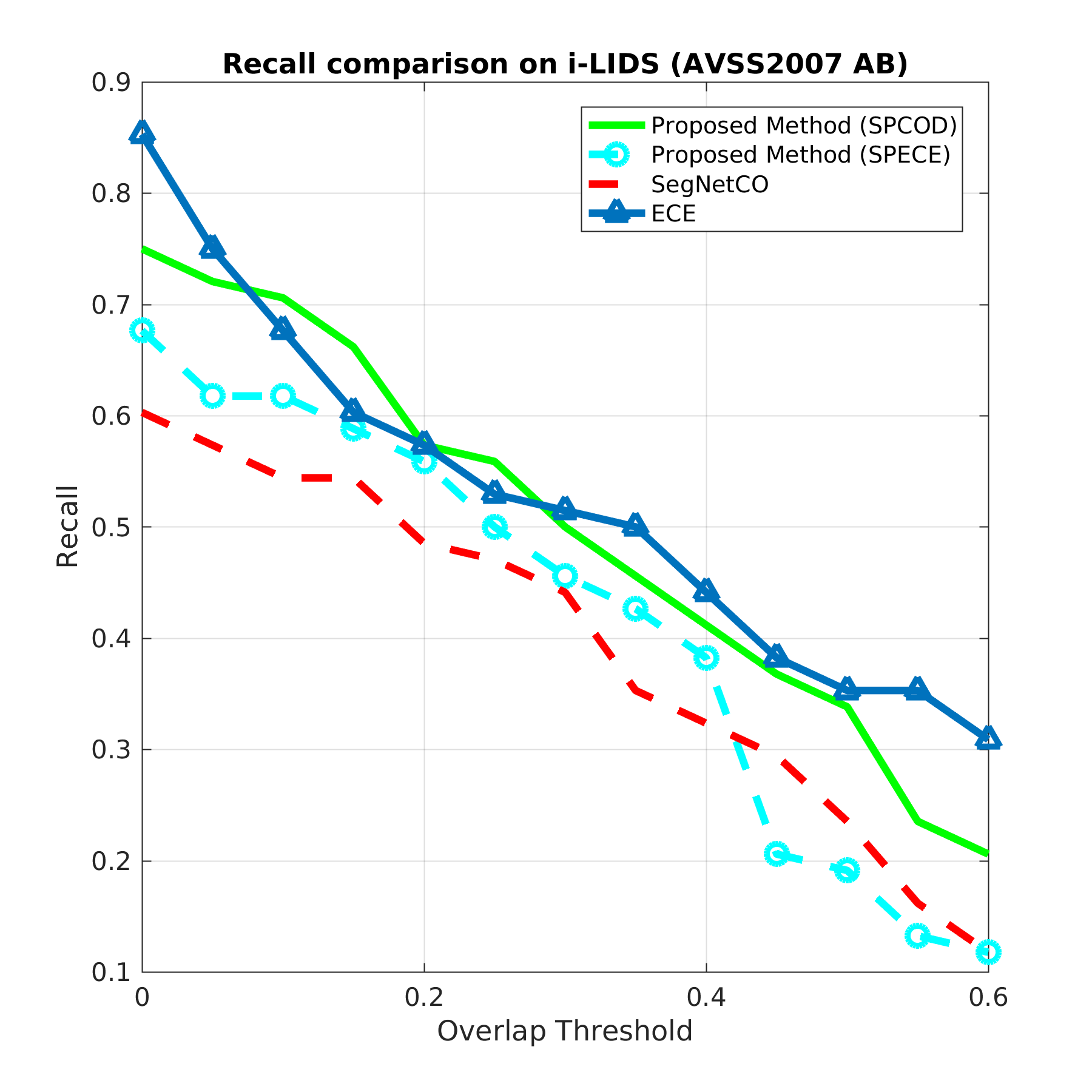}}
\caption{Precision and recall as function of the overlap threshold on i-Lids AVSS.}
\label{A2:PeriLids}
\end{figure*}

\begin{figure*}[htbp]
\captionsetup[subfigure]{labelformat=empty}
\centering
\subfloat[][(c)]{\includegraphics[width=0.48\linewidth]{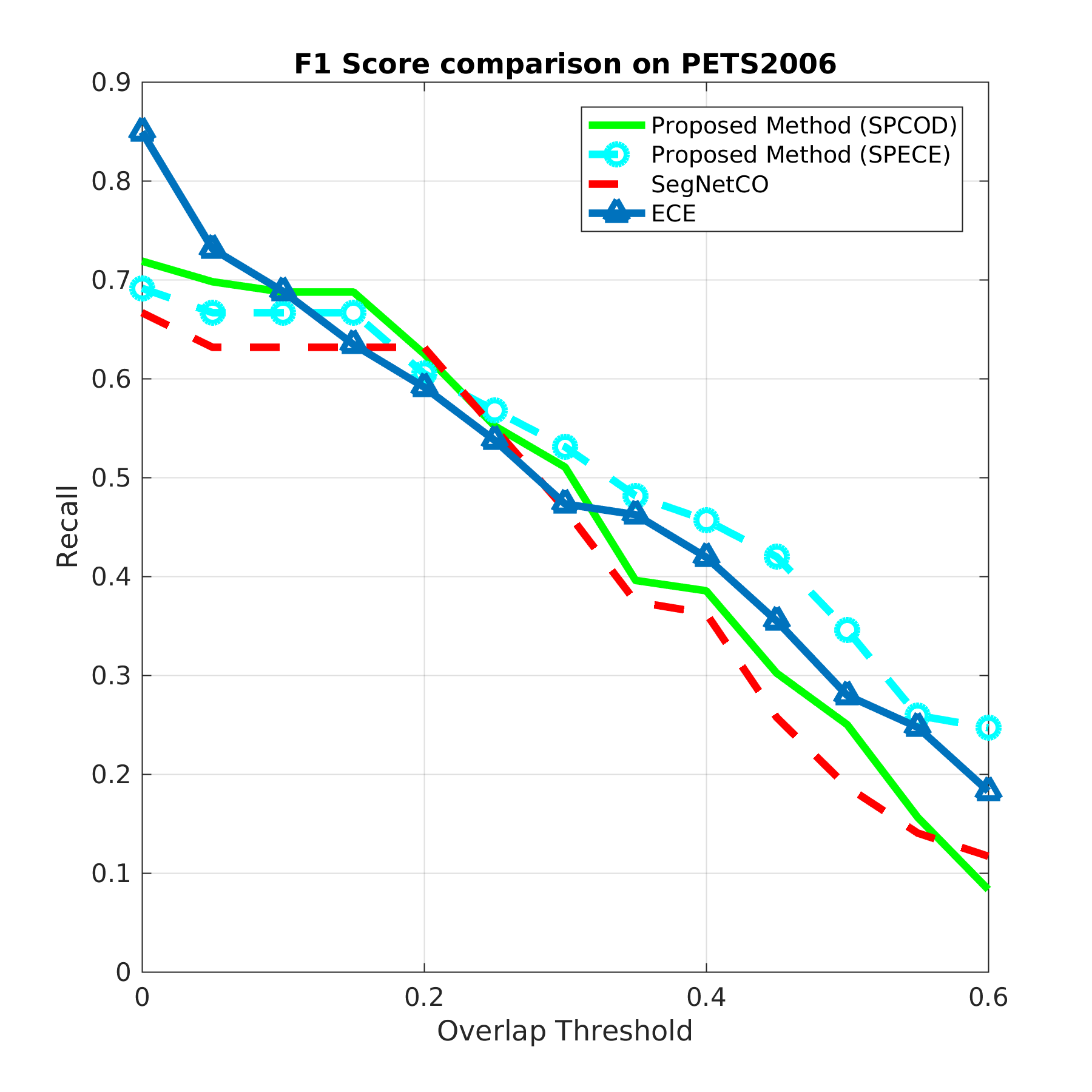}} \hspace{0.06cm}
\subfloat[][(c)]{\includegraphics[width=0.48\linewidth]{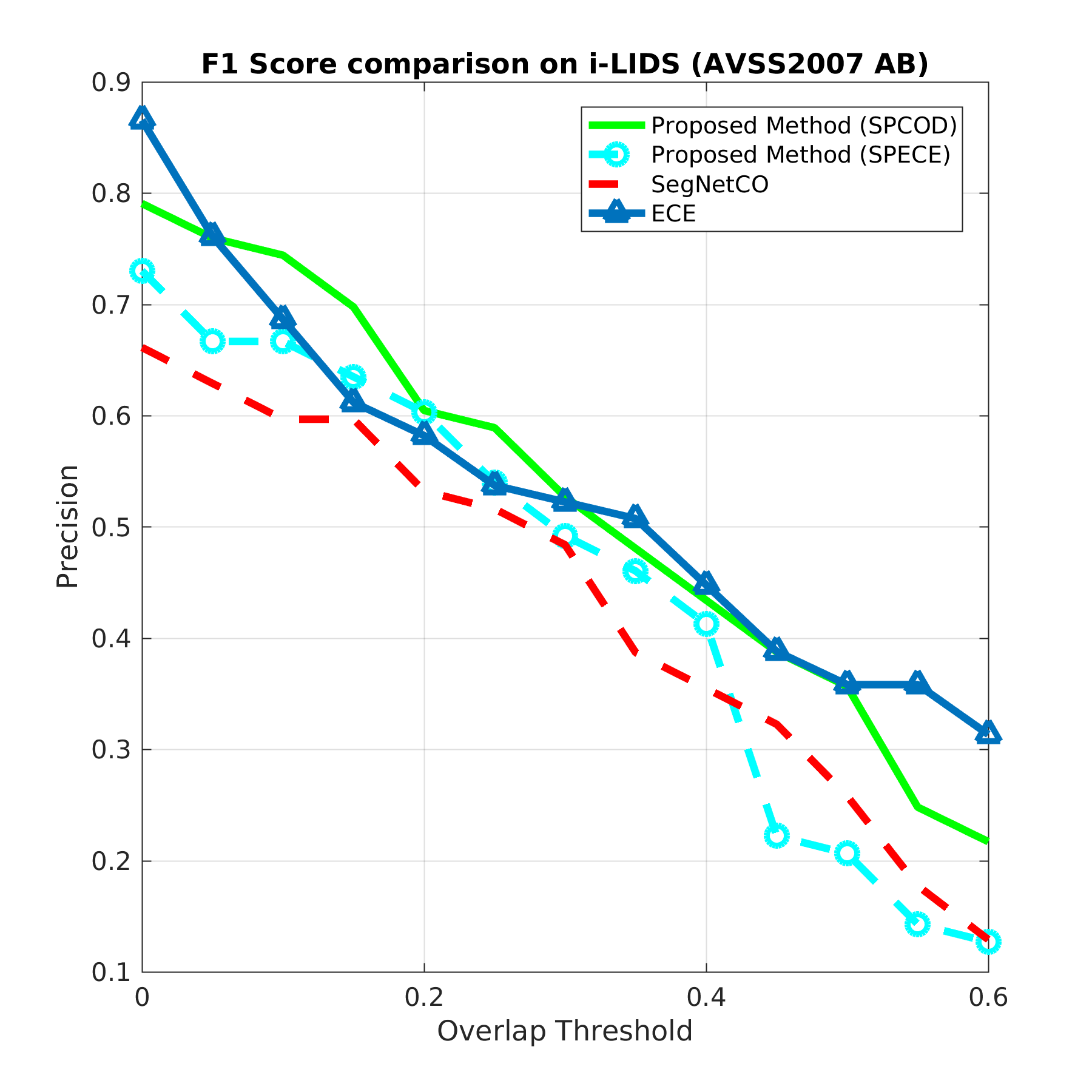}}
\caption{F1 score plots as function of the overlap threshold on i-Lids AVSS.}
\label{A2:F1Lid}
\end{figure*}

\begin{figure}[t]
\captionsetup[subfigure]{labelformat=empty}
     \centering
\begin{tabular}[t]{cccc}
ECE\cite{Ghadiri2016} 
\end{tabular}

\subfloat[][]{\includegraphics[width=1.6cm,height=3cm]{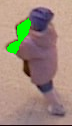}}
\subfloat[][]{\includegraphics[width=1.6cm,height=3cm]{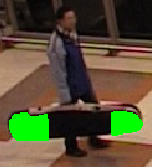}}
\subfloat[][]{\includegraphics[width=1.6cm,height=3cm]{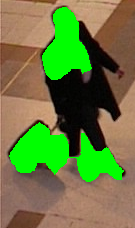}}
\subfloat[][]{\includegraphics[width=1.6cm,height=3cm]{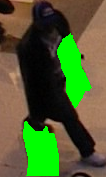}}
\subfloat[][]{\includegraphics[width=1.6cm,height=3cm]{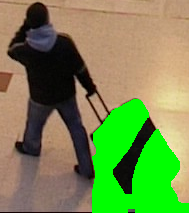}}
\subfloat[][]{\includegraphics[width=1.6cm,height=3cm]{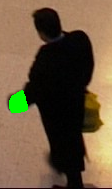}}
\subfloat[][]{\includegraphics[width=1.6cm,height=3cm]{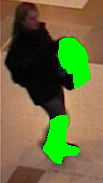}}
\subfloat[][]{\includegraphics[width=1.6cm,height=3cm]{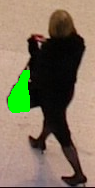}}
\\
     \vspace{-1.9\baselineskip}
\subfloat[][a]{\includegraphics[width=1.6cm,height=3cm]{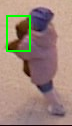}}
\subfloat[][b]{\includegraphics[width=1.6cm,height=3cm]{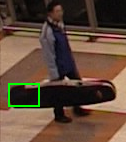}}
\subfloat[][c]{\includegraphics[width=1.6cm,height=3cm]{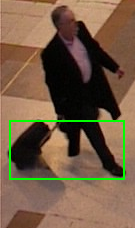}}
\subfloat[][d]{\includegraphics[width=1.6cm,height=3cm]{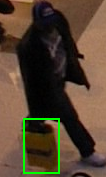}}
\subfloat[][e]{\includegraphics[width=1.6cm,height=3cm]{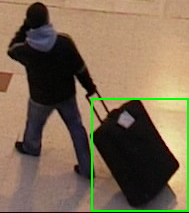}}
\subfloat[][f]{\includegraphics[width=1.6cm,height=3cm]{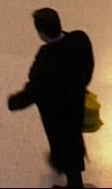}}
\subfloat[][g]{\includegraphics[width=1.6cm,height=3cm]{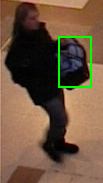}}
\subfloat[][h]{\includegraphics[width=1.6cm,height=3cm]{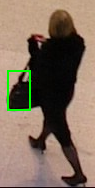}}\\
\begin{tabular}[t]{cccc}
Proposed Method (SPCOD)
\end{tabular}
\subfloat[][]{\includegraphics[width=1.6cm,height=3cm]{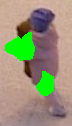}}
\subfloat[][]{\includegraphics[width=1.6cm,height=3cm]{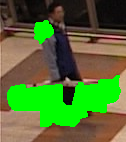}}
\subfloat[][]{\includegraphics[width=1.6cm,height=3cm]{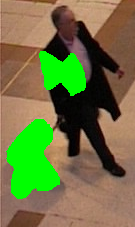}}
\subfloat[][]{\includegraphics[width=1.6cm,height=3cm]{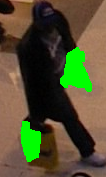}}
\subfloat[][]{\includegraphics[width=1.6cm,height=3cm]{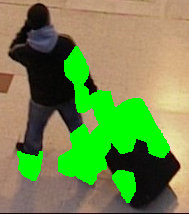}}
\subfloat[][]{\includegraphics[width=1.6cm,height=3cm]{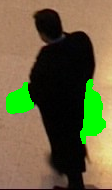}}
\subfloat[][]{\includegraphics[width=1.6cm,height=3cm]{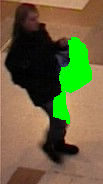}}
\subfloat[][]{\includegraphics[width=1.6cm,height=3cm]{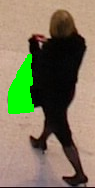}}\\
\vspace{-1.9\baselineskip}
\subfloat[][i]{\includegraphics[width=1.6cm,height=3cm]{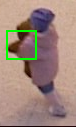}}
\subfloat[][j]{\includegraphics[width=1.6cm,height=3cm]{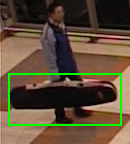}}
\subfloat[][k]{\includegraphics[width=1.6cm,height=3cm]{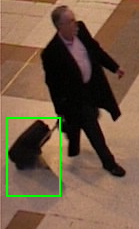}}
\subfloat[][l]{\includegraphics[width=1.6cm,height=3cm]{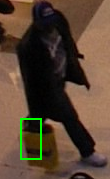}}
\subfloat[][m]{\includegraphics[width=1.6cm,height=3cm]{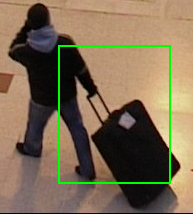}}
\subfloat[][n]{\includegraphics[width=1.6cm,height=3cm]{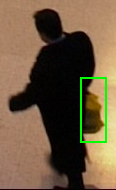}}
\subfloat[][o]{\includegraphics[width=1.6cm,height=3cm]{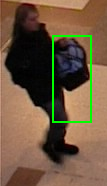}}
\subfloat[][p]{\includegraphics[width=1.6cm,height=3cm]{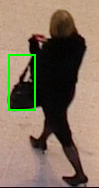}}
     \caption{Successes and failures of our approach on PETS 2006 in comparison with ECE's results. First row: results after segmentation of probable carried objects. Second row: bounding boxes (BB). Green BBs are TP detections, and segmented regions that are not bounded are FP detections. Figure (a-h) results of CO detection obtained by ECE. Figure (i-o) results of our CO detector (SPCOD).}
\label{A2:PETECE}
\end{figure}

\begin{figure}[t]
\captionsetup[subfigure]{labelformat=empty}
     \centering
\begin{tabular}[t]{cccc}
ECE \cite{Ghadiri2016}
\end{tabular}
\vspace{-0.6\baselineskip}
\\
\subfloat[]{\includegraphics[width=1.6cm,height=3cm]{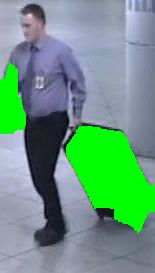}}
\subfloat[]{\includegraphics[width=1.6cm,height=3cm]{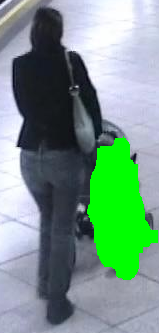}}
\subfloat[]{\includegraphics[width=1.6cm,height=3cm]{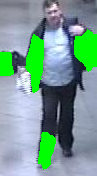}}
\subfloat[]{\includegraphics[width=1.6cm,height=3cm]{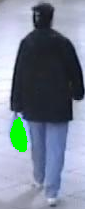}}
\subfloat[]{\includegraphics[width=1.6cm,height=3cm]{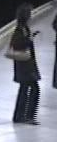}}\\
\vspace{-1.9\baselineskip}
\subfloat[a]{\includegraphics[width=1.6cm,height=3cm]{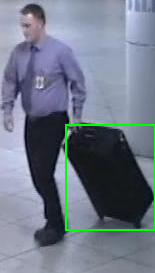}}
\subfloat[b]{\includegraphics[width=1.6cm,height=3cm]{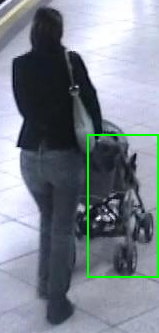}}
\subfloat[c]{\includegraphics[width=1.6cm,height=3cm]{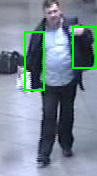}}
\subfloat[d]{\includegraphics[width=1.6cm,height=3cm]{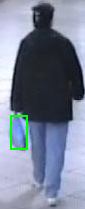}}
\subfloat[e]{\includegraphics[width=1.6cm,height=3cm]{SP/Figure/ILids/ECCV/rec_15.png}}\\
\vspace{-0.6\baselineskip}
\begin{tabular}[t]{cccc}
Proposed Method (SPCOD)
\end{tabular}
\vspace{-0.6\baselineskip}
\\
\subfloat[]{\includegraphics[width=1.6cm,height=3cm]{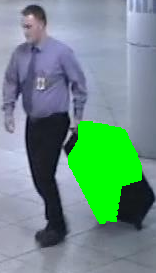}}
\subfloat[]{\includegraphics[width=1.6cm,height=3cm]{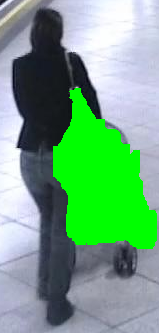}}
\subfloat[]{\includegraphics[width=1.6cm,height=3cm]{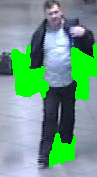}}
\subfloat[]{\includegraphics[width=1.6cm,height=3cm]{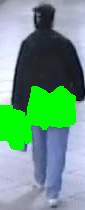}}
\subfloat[]{\includegraphics[width=1.6cm,height=3cm]{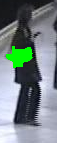}}\\
\vspace{-1.9\baselineskip}
\subfloat[f]{\includegraphics[width=1.6cm,height=3cm]{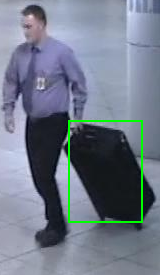}}
\subfloat[g]{\includegraphics[width=1.6cm,height=3cm]{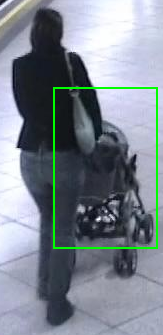}}
\subfloat[h]{\includegraphics[width=1.6cm,height=3cm]{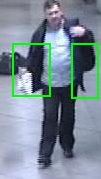}}
\subfloat[i]{\includegraphics[width=1.6cm,height=3cm]{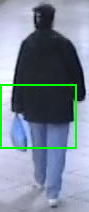}}
\subfloat[j]{\includegraphics[width=1.6cm,height=3cm]{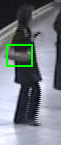}}
    \caption{Successes and failures of our approach on i-Lids AVSS. First row: Segmentation of probable carried object regions. Second row: bounding boxes (BB). Green BBs are TP detections, and segmented regions that are not delineated with a bounding box are detected as FP detections. First row: results obtained by applying ECE method on i-Lids AVSS. Second row: results of our CO detector (SPCOD).}
\label{A2:LidECE}
\end{figure}

We present quantitative comparisons for detection accuracy of four aforementioned algorithms in Tables~\ref{A2:Pet:t} and~\ref{A2:ilidECE}. These tables show the results of our method in comparison to SPECE, ECE and SegNetCO at the overlap threshold value $K=0.15$ (as commonly used in CO detection, see Equation ~\ref{A2:Eval}). To compare our method on i-Lids AVSS and PETS 2006 with ECE, we use their results as stated in their paper. We also illustrate the results as precision, recall and F1 score for increasing values of $K$ in Figures~\ref{A2:Per:P},~\ref{A2:PeriLids} and ~\ref{A2:F1Lid} respectively. Qualitative comparisons are presented in Figures~\ref{A2:PETECE} and~\ref{A2:LidECE}.
\begin{table*}[t]
\begin{center}
\begin{threeparttable}
\begin{tabular}{|l |c |c |c |c |c|c|c|}
\hline
&Prec. & Rec. & TP & FP & FN & F1 Score\\
\hline\hline
Proposed method (SPCOD)  &$60$\%&$\bf79$\%&$\bf66$&$43$&$\bf16$&$\bf 68$\%\\
ECE \cite{Ghadiri2016} &$ 57$\%&$71$\%&$59$&$44$&$24$&$63$\%\\
Variant of proposed method (SPECE)  &$\bf68$\%&$65$\%&$54$&$\bf25$&$29$&$66$\%\\
SegNetCO  &$ 61$\%&$65$\%&$54$&$34$&$29$&$63$\%\\
\hline
\end{tabular}
\end{threeparttable}
\end{center}
\caption{Comparison of our method (SPCOD) with the state-of-the-are methods over PETS 2006 with a $K=0.15$ overlap threshold. The entries with the best value for each evaluation metric are bold-faced.}\label{A2:Pet:t}
\end{table*}

\begin{table*}[t]
\begin{center}
\begin{tabular}{|l |c |c |c |c |c|c|c|}
\hline
&Prec. & Rec. & TP & FP & FN & F1 Score\\
\hline\hline
Proposed method (SPCOD)&\bf 72\%&\bf 64\%&\bf44&\bf17&\bf24&\bf67\%\\
ECE \cite{Ghadiri2016}& 62\%&60\%&41&25&27&60\%\\
Variant of proposed method (SPECE)&68\%&58\%&40&18&24&62\%\\
SegNetCO & 66\%&54\%&37&19&31&59\%\\
\hline
\end{tabular}
\end{center}
\caption{Comparison of our method (SPCOD) with the state-of-the-are methods over i-Lids AVSS with a $K=0.15$ overlap threshold. The entries with the best value for each evaluation metric are bold-faced.}\label{A2:ilidECE}
\end{table*}

As discussed in the following, our experiments demonstrate that our methods (both SPCOD and the SPECE variant) provide the best overall performance in detection of carried objects when the overlap ratio in Equation~\ref{A2:Eval} is between $0.1<K<0.35$. Our algorithm achieves about $4\%$ and $10\%$ gain in performance compared to ECE on PETS 2006 and i-Lids AVSS respectively when $K=0.15$ (Table ~\ref{A2:Pet:t} and Table~\ref{A2:ilidECE}). We also achieve higher recall compared to other methods when the overlap ratio in Equation~\ref{A2:Eval} is between $0.1<K<0.35$. Comparison between SPECE and ECE shows that our new superpixel feature extractor works better than the original one in ECE in terms of F1 score on PETS 2006, while the results on i-Lids AVSS are similar. 

Comparing our method with SegNetCO in figures ~\ref{A2:Per:P} and~\ref{A2:PeriLids}, shows that SegNetCO method have achieved success in terms of precision over SPCOD when the threshold is less than 0.3 in only one dataset (PETS 2006), but it fails when it comes to recall in both datasets. Two main reasons which affect the results obtained by the SegNetCO method are the number of images and quality of the training set. Our training set which is used for SegNetCO method includes high quality images from Fashiniosta dataset where the subject is usually placed leveled with the camera which is different from a video captured by a surveillance camera. Moreover, variety of carried objects in both training set and the set that we used for fine tuning the network is low in comparison to that of in the test set. In general, SegNetCO suffers from the same problems that other learning methods have in case of unseen objects. Objects carried by people can vary significantly in type which makes it particularly challenging to learn models for each object type. Modifying SegNetCO by segmenting human regions (red color in Figure \ref{A2:DeepPIC}(b)), we also investigated if we can have a generic carried object detector by subtracting human regions from foreground mask and then explaining the remaining areas as carried objects. By adding up the results obtained from the person's region subtraction, we gain $2\%$ and $6\%$ on recall in PETS 2006 and i-Lids AVSS respectively when $K=0.15$, but precisions dropped significantly to $55\%$ in PETS 2006 and $59\%$ in i-Lids. This finding shows that segmented human regions by SegNetCO cannot be used in an indirect way for carried object detection and SegNetCO is still far from a generic CO detector.

\begin{figure*}[htbp]
\captionsetup[subfigure]{labelformat=empty}
\centering
\subfloat[]{\includegraphics[width= 2.1cm,height=2.9cm]{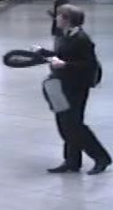}}
\hspace{0.5cm}
\subfloat[]{\includegraphics[width= 2.1cm,height=2.9cm]{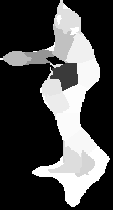}}
\hspace{0.1cm}
\subfloat[]{\includegraphics[width= 2.1cm,height=2.9cm]{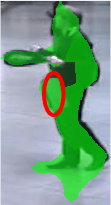}}
\hspace{0.5cm}
\subfloat[]{\includegraphics[width= 2.1cm,height=2.9cm]{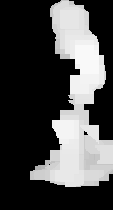}}
\hspace{0.1cm}
\subfloat[]{\includegraphics[width= 2.1cm,height=2.9cm]{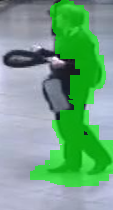}}
\vspace{-1.5\baselineskip}
\\
\subfloat[]{\includegraphics[width= 2.1cm,height=2.9cm]{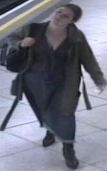}}
 \hspace{0.5cm}
 \subfloat[]{\includegraphics[width= 2.1cm,height=2.9cm]{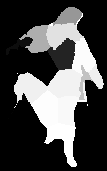}}
 \hspace{0.1cm}
 \subfloat[]{\includegraphics[width= 2.1cm,height=2.9cm]{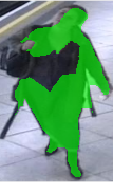}}
 \hspace{0.5cm}
 \subfloat[]{\includegraphics[width= 2.1cm,height=2.9cm]{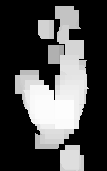}}
 \hspace{0.1cm}
 \subfloat[]{\includegraphics[width= 2.1cm,height=2.9cm]{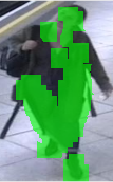}}
 \\
 \vspace{-1.5\baselineskip}
\subfloat[]{\includegraphics[width= 2.1cm,height=2.9cm]{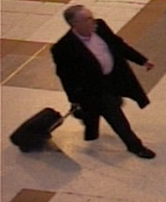}}
\hspace{0.5cm}
\subfloat[a]{\includegraphics[width= 2.1cm,height=2.9cm]{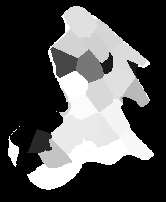}}
\hspace{0.1cm}
\subfloat[b]{\includegraphics[width= 2.1cm,height=2.9cm]{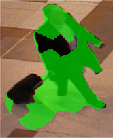}}
 \hspace{0.5cm}
\subfloat[c]{\includegraphics[width= 2.1cm,height=2.9cm]{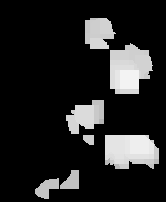}}
\hspace{0.1cm}
\subfloat[d]{\includegraphics[width= 2.1cm,height=2.9cm]{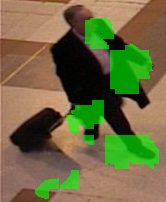}}
\caption{Comparing three person's hypothesis generated from SPECE and ECE. (a, b) person hypothesis by SPECE and its overlay on the original image. (c, d) person hypothesis by ECE and its overlay on the original image.}
\label{A2:CompaerHypo}
\end{figure*}

Comparing SPECE and ECE with regard to precision and recall results shows that although the number of true positives is decreased, our superpixel-based feature extractor can considerably help the system to reduce the number of false positives and as a result, we obtain a better precision with SPECE on both PETS 2006 and i-Lids AVSS. The reason for the decrease in TPs and FPs comes from the generated person hypothesis map by our new feature extractor. Figure~\ref{A2:CompaerHypo} shows three persons with their generated hypothesis map by SPECE and ECE. As it can be seen, the person hypothesis map built by SPECE can better discriminate a person regions from carried object regions. As a result, the number of probable regions for carried objects is reduced which leads to less FPs. However, producing a better hypothesis for the person region may not always lead to better carried object detection. First row in Figure~\ref{A2:CompaerHypo} (a,b) shows that the light grey carried object (delineated by red circle) has a high probability value of being the person region while the hypothesis for the same object built by ECE as shown in the first row of Figure~\ref{A2:CompaerHypo} (c,d) has a very low probability of a being person region. ECE would better detect the object in that case.

As discussed previously, building an accurate person hypothesis plays an important role in finding probable areas for carried objects. In addition to having a well-defined person hypothesis, the way that the remaining areas (probable areas of carried object) are explained in terms of COs is also critical. Here by proposing SPCOD, we show that our superpixel grouping method along with our new feature extractor boost the system recall on both datasets while we increase the overlapping threshold up to $K<0.4$ in comparison to SPECE. Considering qualitative results shown in Figure~\ref{A2:PETECE} and~\ref{A2:LidECE}, SPCOD can detect the carried objects as successfully as ECE. However, our new method can detect smaller objects compared to ECE. SPCOD can successfully detect two small shoulder bags that are completely overlapped with the person's body see Figure ~\ref{A2:LidECE} (g, j) and a small white object carried by the person in Figure~\ref{A2:LidECE} (h) while ECE method fails to detect them (Figure~\ref{A2:LidECE} (b,e,c)). The detection results in Figure~\ref{A2:LidECE} (g,h,i) and Figure~\ref{A2:PETECE} (j) can explain the results in Figures~\ref{A2:PeriLids} and~\ref{A2:Per:P}. As it can be seen, our method successfully detects a higher number of carried objects if $K<0.35$ compared to ECE. However, our method fails to segment carried objects as precisely as in ECE. Therefore, when the overlapping parameter $K$ is increased over $35\%$, detection rate is decreasing. The reason behind this behavior is the quality of the generated hypothesis for a person and the way that superpixel candidates for carried objects are analyzed. The superpixel-wise CO probability map analysis (section~\ref{A2:Contour_assist}) is done by computing informative value of each superpixel in the third superpixel scale level $L_3$. In some cases, the generated superpixels fail to follow object boundaries and the information obtained from boundary support and shape context descriptors is not rich enough to differentiate between superpixels belonging to a carried object and a person.   

\subsection{Ablation Study}
In this section, we conducted an ablation study to understand how critical are the feature extraction and person viewpoint (VP) detection for the performance of the carried object detection. First, we disabled the functionality of viewpoint detection and performed experiments on both i-Lids AVSS and PETS 2006 datasets. The precision on i-Lids AVSS drops to $46\%$ and precision on PETS 2006 drops to $40\%$ when $K=0.15$ (Table ~\ref{A2:Ablation}). This finding suggests that detecting carried objects strongly rely on choosing right viewpoint exemplars.

\begin{table}
\centering 
\begin{tabular}{|l |c |c |c |c |c|c|c|}
\hline
\bf PETS 2006&Prec. & Rec. &F1 score \\
\hline\hline
Proposed Method(SPCOD)&\bf 60\%&\bf 79\%&\bf 68\%\\
SPCOD without VP detection&40\%&53\%&45\%\\
SPCOD without $L_3$& 43\%&45\%&43\%\\
SPCOD without $L_2$& 41\%&39\%&39\%\\
SPCOD without $L_1$& 36\%&37\%&31\%\\
\hline
\end{tabular}
\caption{Study of the different components of our algorithm on PETS 2006}\label{A2:Ablation}
\end{table}

\begin{table}
\centering 
\begin{tabular}{|l |c |c |c |c |c|c|c|}
\hline
\bf i-Lids AVSS&Prec. & Rec. &F1 score\\
\hline\hline
Proposed Method(SPCOD)&\bf 72\%&\bf 64\%&\bf 64\%\\
SPCOD without VP detection&46\%&41\%&43\%\\
SPCOD without $L_3$& 50\%&43\%&46\%\\
SPCOD without $L_2$& 49\%&41\%&44\%\\
SPCOD without $L_1$& 49\%&41\%&44\%\\
\hline
\end{tabular}
\caption{Study of the different components of our algorithm on i-Lids AVSS}\label{A2:Ablation2}
\end{table}

Next we study the impact of the multiscale superpixel segmentation to extract features. When one of the superpixel level is disabled, precision drops from $60\%$ to $43\%$ on PETS 2006 and it drops to around $50\%$ for i-Lids AVSS. This means that all features extracted from the three superpixel levels contribute to the performance (Table ~\ref{A2:Ablation2}).
 
\section{Conclusion}
\label{A2:conclude}
We presented a method that uses the advantages of learning a codebook of human exemplars to differentiate carried object regions from person regions. We formulated the problem of detecting carried objects as the problem of searching for a subset of superpixels that belong to a person and explaining the remaining superpixels in the foreground mask in terms of carried objects. The proposed algorithm achieves state-of-the-art performance in terms of carried object detection on PETS 2006 and i-Lids AVSS dataset on a large range of the overlap threshold. We also showed that our feature extractor can predict a person hypothesis better than the one in previous work (ECE method). As a result, when we use our feature extractor instead of the one in ECE, the number of false positives decreases dramatically.

\section*{Acknowledgment}
This research was supported by FRQ-NT team grant No. 2014-PR-172083.

\section*{References}

\bibliography{IEEEabrv,mybibfile}

\begin{thebibliography}{10}
\expandafter\ifx\csname url\endcsname\relax
  \def\url#1{\texttt{#1}}\fi
\expandafter\ifx\csname urlprefix\endcsname\relax\def\urlprefix{URL }\fi
\expandafter\ifx\csname href\endcsname\relax
  \def\href#1#2{#2} \def\path#1{#1}\fi

\bibitem{Branca}
A.~Branca, M.~Leo, G.~Attolico, A.~Distante, Detection of objects carried by
  people, in: Image Processing. 2002. Proceedings. 2002 International
  Conference on, Vol.~3, 2002, pp. III--317--III--320 vol.3.
\newblock \href {http://dx.doi.org/10.1109/ICIP.2002.1038969}
  {\path{doi:10.1109/ICIP.2002.1038969}}.

\bibitem{Dondera}
R.~Dondera, V.~Morariu, L.~Davis, Learning to detect carried objects with
  minimal supervision, in: Computer Vision and Pattern Recognition Workshops
  (CVPRW), 2013 IEEE Conference on, 2013, pp. 759--766.
\newblock \href {http://dx.doi.org/10.1109/CVPRW.2013.114}
  {\path{doi:10.1109/CVPRW.2013.114}}.

\bibitem{Damen}
D.~Damen, D.~Hogg, Detecting carried objects from sequences of walking
  pedestrians, Pattern Analysis and Machine Intelligence, IEEE Transactions on
  34~(6) (2012) 1056--1067.
\newblock \href {http://dx.doi.org/10.1109/TPAMI.2011.205}
  {\path{doi:10.1109/TPAMI.2011.205}}.

\bibitem{Haritaoglu}
I.~Haritaoglu, R.~Cutler, D.~Harwood, L.~Davis, Backpack: detection of people
  carrying objects using silhouettes, in: Computer Vision, 1999. The
  Proceedings of the Seventh IEEE International Conference on, Vol.~1, 1999,
  pp. 102--107 vol.1.

\bibitem{Chayanurak}
R.~Chayanurak, N.~Cooharojananone, S.~Satoh, R.~Lipikorn, Carried object
  detection using star skeleton with adaptive centroid and time series graph,
  in: Signal Processing (ICSP), 2010 IEEE 10th International Conference on,
  2010, pp. 736--739.
\newblock \href {http://dx.doi.org/10.1109/ICOSP.2010.5655765}
  {\path{doi:10.1109/ICOSP.2010.5655765}}.

\bibitem{Ghadiri2016}
F.~Ghadiri, R.~Bergevin, G.-A. Bilodeau, Carried object detection based on an
  ensemble of contour exemplars, in: Proceedings of the 14th European
  Conference on Computer Vision-Part VII, Springer International Publishing,
  2016, pp. 852--866.
\newblock \href {http://dx.doi.org/10.1007/978-3-319-46478-7_52}
  {\path{doi:10.1007/978-3-319-46478-7_52}}.

\bibitem{Belongie}
S.~Belongie, J.~Malik, J.~Puzicha, Shape matching and object recognition using
  shape contexts, Pattern Analysis and Machine Intelligence, IEEE Transactions
  on 24~(4) (2002) 509--522.
\newblock \href {http://dx.doi.org/10.1109/34.993558}
  {\path{doi:10.1109/34.993558}}.

\bibitem{JavedOmar}
O.~Javed, M.~Shah, Tracking and object classification for automated
  surveillance, in: A.~Heyden, G.~Sparr, M.~Nielsen, P.~Johansen (Eds.),
  Computer Vision — ECCV 2002, Vol. 2353 of Lecture Notes in Computer
  Science, Springer Berlin Heidelberg, 2002, pp. 343--357.

\bibitem{Lee2006}
C.-S. Lee, A.~Elgammal, Carrying object detection using pose preserving dynamic
  shape models, in: Proceedings of the 4th International Conference on
  Articulated Motion and Deformable Objects, AMDO'06, Springer-Verlag, Berlin,
  Heidelberg, 2006, pp. 315--325.
\newblock \href {http://dx.doi.org/10.1007/11789239_33}
  {\path{doi:10.1007/11789239_33}}.

\bibitem{Tzanidou}
G.~Tzanidou, I.~Zafar, E.~Edirisinghe, Carried object detection in videos using
  color information, Information Forensics and Security, IEEE Transactions on
  8~(10) (2013) 1620--1631.
\newblock \href {http://dx.doi.org/10.1109/TIFS.2013.2279797}
  {\path{doi:10.1109/TIFS.2013.2279797}}.

\bibitem{Tavanai}
A.~Tavanai, M.~Sridhar, F.~Gu, A.~Cohn, D.~Hogg, Carried object detection and
  tracking using geometric shape models and spatio-temporal consistency, in:
  M.~Chen, B.~Leibe, B.~Neumann (Eds.), Computer Vision Systems, Vol. 7963 of
  Lecture Notes in Computer Science, Springer Berlin Heidelberg, 2013, pp.
  223--233.

\bibitem{Senst2012}
T.~Senst, A.~Kuhn, H.~Theisel, T.~Sikora, Detecting people carrying objects
  utilizing lagrangian dynamics, in: Advanced Video and Signal-Based
  Surveillance (AVSS), 2012 IEEE Ninth International Conference on, 2012, pp.
  398--403.

\bibitem{Chua}
T.~W. Chua, K.~Leman, H.~L. Wang, N.~T. Pham, R.~Chang, D.~D. Nguyen, J.~Zhang,
  Sling bag and backpack detection for human appearance semantic in vision
  system, in: 2013 IEEE/RSJ International Conference on Intelligent Robots and
  Systems, 2013, pp. 2130--2135.
\newblock \href {http://dx.doi.org/10.1109/IROS.2013.6696654}
  {\path{doi:10.1109/IROS.2013.6696654}}.

\bibitem{Yue}
Y.~Qi, G.-C. Huang, Y.-H. Wang, Carrying object detection and tracking based on
  body main axis, in: 2007 International Conference on Wavelet Analysis and
  Pattern Recognition, Vol.~3, 2007, pp. 1237--1240.
\newblock \href {http://dx.doi.org/10.1109/ICWAPR.2007.4421623}
  {\path{doi:10.1109/ICWAPR.2007.4421623}}.

\bibitem{DBLP}
K.~He, G.~Gkioxari, P.~Doll{\'{a}}r, R.~B. Girshick,
  \href{http://arxiv.org/abs/1703.06870}{Mask {R-CNN}}, CoRR abs/1703.06870.
\newline\urlprefix\url{http://arxiv.org/abs/1703.06870}

\bibitem{Parsing}
X.~Liang, C.~Xu, X.~Shen, J.~Yang, J.~Tang, L.~Lin, S.~Yan, Human parsing with
  contextualized convolutional neural network, IEEE Transactions on Pattern
  Analysis and Machine Intelligence 39~(1) (2017) 115--127.
\newblock \href {http://dx.doi.org/10.1109/TPAMI.2016.2537339}
  {\path{doi:10.1109/TPAMI.2016.2537339}}.

\bibitem{SegNet}
V.~Badrinarayanan, A.~Kendall, R.~Cipolla, Segnet: A deep convolutional
  encoder-decoder architecture for image segmentation, IEEE Transactions on
  Pattern Analysis and Machine Intelligence.

\bibitem{FRCNN}
S.~Ren, K.~He, R.~Girshick, J.~Sun,
  \href{http://dl.acm.org/citation.cfm?id=2969239.2969250}{Faster r-cnn:
  Towards real-time object detection with region proposal networks}, in:
  Proceedings of the 28th International Conference on Neural Information
  Processing Systems - Volume 1, NIPS'15, MIT Press, Cambridge, MA, USA, 2015,
  pp. 91--99.
\newline\urlprefix\url{http://dl.acm.org/citation.cfm?id=2969239.2969250}

\bibitem{FCN}
E.~Shelhamer, J.~Long, T.~Darrell,
  \href{https://doi.org/10.1109/TPAMI.2016.2572683}{Fully convolutional
  networks for semantic segmentation}, IEEE Trans. Pattern Anal. Mach. Intell.
  39~(4) (2017) 640--651.
\newblock \href {http://dx.doi.org/10.1109/TPAMI.2016.2572683}
  {\path{doi:10.1109/TPAMI.2016.2572683}}.
\newline\urlprefix\url{https://doi.org/10.1109/TPAMI.2016.2572683}

\bibitem{Yamaguchi}
K.~Yamaguchi, M.~H. Kiapour, T.~L. Berg, Paper doll parsing: Retrieving similar
  styles to parse clothing items, in: 2013 IEEE International Conference on
  Computer Vision, 2013, pp. 3519--3526.
\newblock \href {http://dx.doi.org/10.1109/ICCV.2013.437}
  {\path{doi:10.1109/ICCV.2013.437}}.

\bibitem{Fashionista}
K.~Yamaguchi, M.~H. Kiapour, T.~L. Berg, Paper doll parsing: Retrieving similar
  styles to parse clothing items, in: 2013 IEEE International Conference on
  Computer Vision, 2013, pp. 3519--3526.
\newblock \href {http://dx.doi.org/10.1109/ICCV.2013.437}
  {\path{doi:10.1109/ICCV.2013.437}}.

\bibitem{Lin2014}
T.-Y. Lin, M.~Maire, S.~Belongie, J.~Hays, P.~Perona, D.~Ramanan,
  P.~Doll{\'a}r, C.~L. Zitnick,
  \href{https://doi.org/10.1007/978-3-319-10602-1_48}{Microsoft COCO: Common
  Objects in Context}, Springer International Publishing, Cham, 2014, pp.
  740--755.
\newblock \href {http://dx.doi.org/10.1007/978-3-319-10602-1_48}
  {\path{doi:10.1007/978-3-319-10602-1_48}}.
\newline\urlprefix\url{https://doi.org/10.1007/978-3-319-10602-1_48}

\bibitem{Abdelkader}
C.~B. Abdelkader, Detection of people carrying objects: A motion-based
  recognition approach, in: Proceedings of the Fifth IEEE International
  Conference on Automatic Face and Gesture Recognition, FGR '02, IEEE Computer
  Society, Washington, DC, USA, 2002, pp. 378--.

\bibitem{Delgado}
B.~Delgado, K.~Tahboub, E.~J. Delp, Superpixels shape analysis for carried
  object detection, in: 2016 IEEE Winter Applications of Computer Vision
  Workshops (WACVW), 2016, pp. 1--6.
\newblock \href {http://dx.doi.org/10.1109/WACVW.2016.7470116}
  {\path{doi:10.1109/WACVW.2016.7470116}}.

\bibitem{Sander}
J.~Sander, M.~Ester, H.-P. Kriegel, X.~Xu, Density-based clustering in spatial
  databases: The algorithm gdbscan and its applications, Data Mining and
  Knowledge Discovery 2~(2) (1998) 169--194.
\newblock \href {http://dx.doi.org/10.1023/A:1009745219419}
  {\path{doi:10.1023/A:1009745219419}}.

\bibitem{Wahyono2017}
Wahyono, K.-H. Jo, Human Carrying Baggage Classification Using Transfer
  Learning on CNN with Direction Attribute, Springer International Publishing,
  Cham, 2017, pp. 717--724.
\newblock \href {http://dx.doi.org/10.1007/978-3-319-63309-1_63}
  {\path{doi:10.1007/978-3-319-63309-1_63}}.

\bibitem{Ghadiri2017}
F.~Ghadiri, R.~Bergevin, G.-A. Bilodeau, Spatio-temporal consistency to detect
  and segment carried objects, in: Proceedings of the 14th European Conference
  on Computer Vision-Part VII, Springer Berlin Heidelberg, In Prec., 2017.

\bibitem{DPBM}
P.~F. Felzenszwalb, R.~B. Girshick, D.~McAllester, D.~Ramanan, Object detection
  with discriminatively trained part based models, IEEE Transactions on Pattern
  Analysis and Machine Intelligence 32~(9) (2010) 1627--1645.

\bibitem{Charles}
P.-L. St-Charles, G.-A. Bilodeau, R.~Bergevin, A self-adjusting approach to
  change detection based on background word consensus, in: Applications of
  Computer Vision (WACV), 2015 IEEE Winter Conference on, 2015, pp. 990--997.
\newblock \href {http://dx.doi.org/10.1109/WACV.2015.137}
  {\path{doi:10.1109/WACV.2015.137}}.

\bibitem{Achanta}
R.~Achanta, A.~Shaji, K.~Smith, A.~Lucchi, P.~Fua, S.~Susstrunk, Slic
  superpixels compared to state-of-the-art superpixel methods, IEEE Trans.
  Pattern Anal. Mach. Intell. 34~(11) (2012) 2274--2282.
\newblock \href {http://dx.doi.org/10.1109/TPAMI.2012.120}
  {\path{doi:10.1109/TPAMI.2012.120}}.

\bibitem{Martin}
D.~R. Martin, C.~C. Fowlkes, J.~Malik, Learning to detect natural image
  boundaries using local brightness, color, and texture cues, IEEE Trans.
  Pattern Anal. Mach. Intell. 26~(5) (2004) 530--549.
\newblock \href {http://dx.doi.org/10.1109/TPAMI.2004.1273918}
  {\path{doi:10.1109/TPAMI.2004.1273918}}.

\bibitem{Wang}
L.~Wang, J.~Shi, G.~Song, I.-F. Shen, Object detection combining recognition
  and segmentation, in: Proceedings of the 8th Asian Conference on Computer
  Vision - Volume Part I, ACCV'07, Springer-Verlag, Berlin, Heidelberg, 2007,
  pp. 189--199.

\bibitem{Triggs}
N.~Dalal, B.~Triggs, Histograms of oriented gradients for human detection, in:
  Computer Vision and Pattern Recognition, 2005. CVPR 2005. IEEE Computer
  Society Conference on, Vol.~1, 2005, pp. 886--893 vol. 1.
\newblock \href {http://dx.doi.org/10.1109/CVPR.2005.177}
  {\path{doi:10.1109/CVPR.2005.177}}.

\end{thebibliography}

\end{document}